\documentclass[11pt]{article}

\usepackage[preprint]{acl}

\usepackage{times}
\usepackage{latexsym}
\usepackage{booktabs}
\usepackage{multirow}
\usepackage[T1]{fontenc}
\usepackage[utf8]{inputenc}

\usepackage{microtype}

\usepackage{inconsolata}

\usepackage{graphicx}
\usepackage{amsmath}
\usepackage{amssymb} % for \mathbb
\usepackage{booktabs}
\usepackage{tabularx}
\usepackage{mathtools}
\usepackage{colortbl} % must be loaded before arydshln
\usepackage{arydshln}
\usepackage{multirow} 
\usepackage{stfloats}
\usepackage[figuresright]{rotating}
\usepackage{float}
\usepackage{placeins}
\usepackage{url}
\usepackage{listings}
\usepackage[ruled,vlined,linesnumbered]{algorithm2e}
\usepackage{xcolor}
\usepackage{CJKutf8}
\SetKwProg{Procedure}{procedure}{}{end}
\SetKw{Return}{return}
\SetKwComment{Comment}{$\triangleright$\ }{}

\newcommand{\zhname}[1]{\begin{CJK}{UTF8}{gbsn}#1\end{CJK}}

\def\code#1{\texttt{#1}}
\title{Quit While You're Ahead: \textsc{Quit} for Efficient Candidate Generation in Machine Translation Reranking}

\author{\textbf{Guangyu Chen}\thanks{Equal contribution.}$^{\text{1}}$, \textbf{Boxuan Lyu \zhname{吕博轩}}\footnotemark[1]$^{\text{1}}$, \\ 
 \textbf{Hidetaka Kamigaito}$^{\text{2}}$, \textbf{Kotaro Funakoshi}$^{\text{1}}$, and \textbf{Manabu Okumura}$^{\text{1}}$ \vspace{1mm}\\ 
  $^{\text{1}}$Institute of Science Tokyo,  $^{\text{2}}$Nara Institute of Science and Technology\\
  \texttt{\url{{co_gy,lyu,funakoshi,oku}@lr.first.iir.isct.ac.jp}}\\ \texttt{\url{kamigaito.h@is.naist.jp}}}

\begin{document}
\maketitle
\begin{abstract}
Reranking methods, such as Minimum Bayes Risk (MBR) decoding and Quality Estimation (QE) reranking, have been widely used in modern neural machine translation (NMT) to select an output from a set of candidate hypotheses. 
However, the performance gains come at the cost of high inference latency.
Existing acceleration methods target MBR decoding and reduce only the reranking computation, leaving QE reranking unaddressed and candidate generation---which can be the larger computational bottleneck---largely untouched.
In this work, we propose \textbf{\textsc{Quit}} (\textbf{Q}uantifying \textbf{U}ncertainty for \textbf{I}ncremental \textbf{T}ermination), a novel early-stopping strategy for the entire generation--reranking pipeline.
\textsc{Quit} treats candidate generation as a sequential decision-making process under uncertainty.
It incrementally generates and reranks candidates, stopping when the best reranking score stabilizes.
Comprehensive experiments with three NMT models across 19 language pairs show that \textsc{Quit} achieves end-to-end speedups of $1.47$--$2.66\times$ for MBR decoding and $3.43$--$4.12\times$ for QE reranking while preserving automatic metric scores. 
%Code is available at \url{https://anonymous.4open.science/r/QUIT-code}. 
\end{abstract}

\section{Introduction}

Reranking is increasingly common in modern neural machine translation (NMT): a reranker selects the final output from a set of hypotheses generated by an NMT model. 
Reranking offers two advantages: (1) it uses external criteria rather than relying solely on model probability, which has been shown to correlate poorly with translation quality \cite{is_map_need,reranking_nerual_metric} and (2) it flexibly adapts output selection to different inference-time objectives without requiring model retraining \cite{gender_reranking}. 
In recent WMT General Translation Shared Tasks, nearly all top-performing systems have used reranking during inference or for training-data construction \cite{wmt24,wmt25}. 
The predominant approaches are Minimum Bayes Risk (MBR) decoding \cite{smt_mbr1,smt_mbr2,is_map_need,nmt_mbr_surface2,reranking_nerual_metric,qa_reranking} and Quality Estimation (QE) reranking \cite{qa_reranking}, along with their variants.

Despite its effectiveness, the reranking pipeline is computationally expensive, as it requires generating many hypotheses and repeatedly invoking the reranker. 
Although several methods have recently been proposed to accelerate MBR decoding \cite{CPMBR,RefAggMBR,CentroidMBR}, they are specific to MBR and optimize only the reranking stage, without reducing the need for large candidate sets. 
To the best of our knowledge, no acceleration method has been developed for QE reranking. 
Moreover, as large language model (LLM)-based NMT systems become more common, candidate generation can be substantially more costly than reranking.
Therefore, reducing the number of generated candidates while maintaining the final translation quality---as measured against references---holds great promise for substantial end-to-end acceleration.

Viewing the reranking pipeline as an incremental generation--reranking process, an intuitive idea is to terminate generation early when the risk is minimal, that is, when producing additional candidates is unlikely to yield further quality improvements. 
This early-stopping risk can be characterized as the absolute difference in quality between the reranker's current output and its output from the full candidate set.
However, in an incremental setting, the output quality of the full candidate set is inaccessible. 
We thus utilize uncertainty, quantified as the degree of quality convergence throughout the incremental process, as a proxy for this risk.
Furthermore, because true translation quality can only be measured against references (which are unavailable at test time), we leverage the score assigned by the reranker itself as a practical surrogate.
We operationalize this framework into \textbf{\textsc{Quit}} (\textbf{Q}uantifying \textbf{U}ncertainty for \textbf{I}ncremental \textbf{T}ermination), an uncertainty-guided early-stopping strategy for the entire generation--reranking pipeline.
Concretely, \textsc{Quit} generates candidates incrementally (e.g., eight at a time) and tracks their reranking scores. It uses the recent variation of the best reranking score as a practical proxy for this uncertainty and stops when that variation is at or below a prespecified convergence threshold.
This strategy therefore reduces both candidate-generation and reranking costs.

\begin{figure*}
    \centering
    \includegraphics[width=1\linewidth]{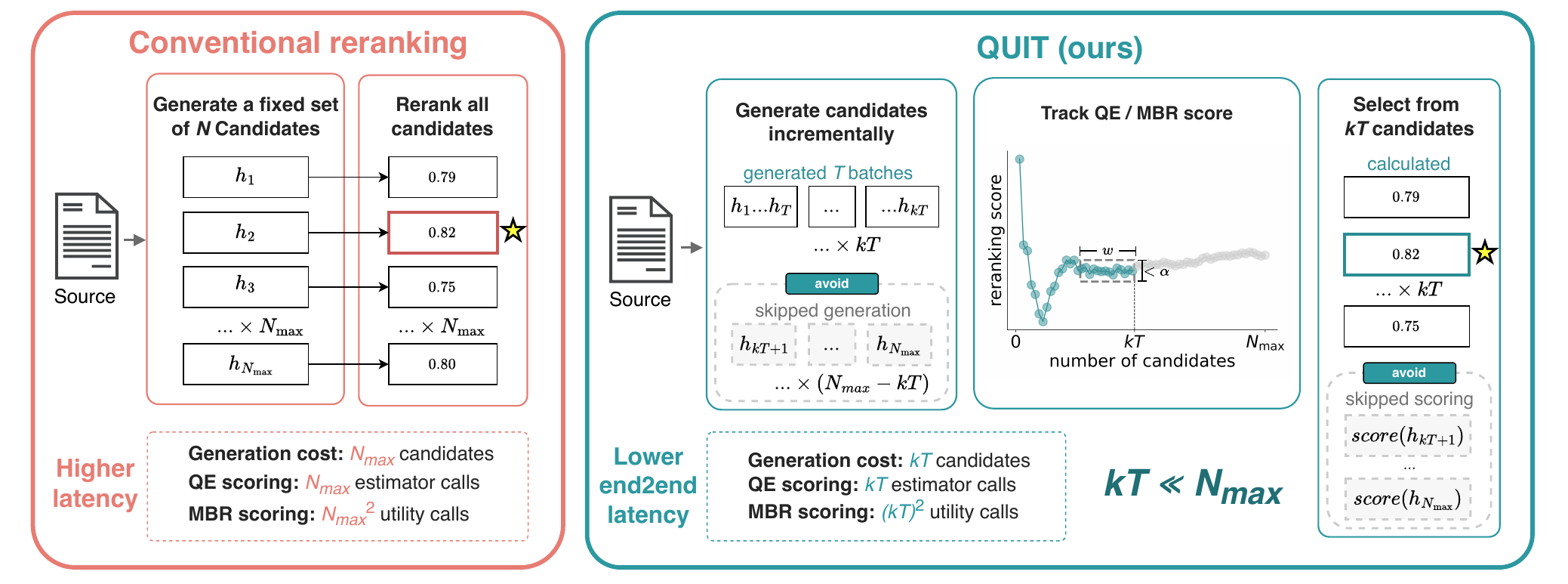}
    \caption{Overview of conventional reranking and \textsc{Quit}. Conventional reranking generates the full candidate set and scores every candidate, whereas \textsc{Quit} generates candidates incrementally, tracks the best reranking score, and stops once that score stabilizes. It thereby avoids generating and scoring additional candidates and reduces end-to-end latency.}
    \label{fig:placeholder}
\end{figure*}
Across three NMT models and the WMT24 \cite{wmt24} and WMT25 test sets, spanning 19 language pairs, \textsc{Quit} achieves end-to-end speedups of $1.47$--$2.66\times$ for MBR decoding and $3.43$--$4.12\times$ for QE reranking while remaining statistically equivalent to the unaccelerated baseline for most automatic metric scores.
These results demonstrate that adaptive candidate generation is key to reducing end-to-end reranking latency.

\section{NMT Reranking}
\label{sec:nmt_reranking}
Reranking selects a translation according to the preferences of an external reranker, rather than solely according to the probabilities estimated by the NMT model.
An NMT system with reranking consists of two stages.
First, the NMT model generates multiple candidates, for example, through beam search or sampling.
Second, the reranker scores these candidates, and the system returns the one with the best reranking score.
Decoupling selection from generation allows the output selection to target criteria that may align better with translation quality than model probability.

Formally, given a source sentence $s$, an NMT model with parameters $\theta$ defines a distribution $P_{\theta}(\cdot \mid s)$ over the hypothesis space $\mathcal{H}$.
A generation method constructs a candidate set $\mathcal{C}=\{h_1,\ldots,h_N\}\subset\mathcal{H}$ of size $N$.
Let $r(h;s,\mathcal{C})$ denote the score that an external reranker assigns to candidate $h\in\mathcal{C}$, where a higher score indicates a stronger preference.
The system selects:
\begin{equation}
\hat{y}_r = \mathop{\mathrm{argmax}}\limits_{h\in\mathcal{C}}
r(h;s,\mathcal{C}).
\label{eq:reranking}
\end{equation}
The resulting performance depends on both the coverage of $\mathcal{C}$ and the accuracy of $r$.
Increasing $N$ makes a high-quality translation more likely to appear in $\mathcal{C}$, but also raises the cost of both stages.

\subsection{Minimum Bayes Risk Decoding}
\label{sec:mbr}
MBR decoding is a reranking method with a long history in statistical machine translation \cite{smt_mbr1,smt_mbr2}.
It uses hypotheses sampled from the model as ``pseudo-references'', called \emph{support hypotheses}, and scores each candidate by its average utility against them.
The candidate with the lowest estimated Bayes risk---or, equivalently, the highest expected utility---is returned as the final translation.

Formally, MBR instantiates the generic score $r$ in Equation~\ref{eq:reranking} as an expected-utility score $r_{\mathrm{MBR}}$.
Let $\mathcal{S}=\{h_s^{(m)}\}_{m=1}^{M}$ denote a set of $M$ independently sampled support hypotheses, where $h_s^{(m)}\sim P_{\theta}(\cdot\mid s)$.
Let $u(h,h_s)$ denote the utility of candidate $h$ when support hypothesis $h_s\in\mathcal{S}$ serves as a pseudo-reference.
The Monte Carlo MBR score and the corresponding selected output are:
\begin{align}
r_{\mathrm{MBR}}(h;\mathcal{S})
&= \frac{1}{|\mathcal{S}|}
\sum_{h_s\in\mathcal{S}} u(h,h_s), \\
\hat{y}_{\mathrm{MBR}}
&= \mathop{\mathrm{argmax}}\limits_{h\in\mathcal{C}}
r_{\mathrm{MBR}}(h;\mathcal{S}).
\label{eq:mbr}
\end{align}
This uniform average is the standard Monte Carlo approximation to the expectation under the model distribution.
In standard MBR decoding, the candidate and support sets are often identical, i.e., $\mathcal{C}=\mathcal{S}$.
Neural evaluation metrics can serve as $u$, allowing selection to target translation quality more directly than model probability does.
This benefit, however, comes at a substantial computational cost: scoring all candidate--support pairs requires $|\mathcal{C}|\times|\mathcal{S}|$ utility evaluations and becomes quadratic in the candidate-set size when $\mathcal{C}=\mathcal{S}$.

\subsection{Quality Estimation Reranking}
\label{sec:qe_reranking}

QE reranking dispenses with support hypotheses entirely.
Instead, a QE model predicts translation quality directly from the source sentence and a candidate, without access to a human reference translation.

Formally, QE instantiates $r$ as a reference-free score $r_{\mathrm{QE}}$.
This score represents the predicted quality of candidate $h$ given the source sentence $s$ and, unlike the MBR score, does not depend on the other candidates in $\mathcal{C}$.
The QE-selected output is:
\begin{equation}
\hat{y}_{\mathrm{QE}}
= \mathop{\mathrm{argmax}}\limits_{h\in\mathcal{C}}
r_{\mathrm{QE}}(h;s).
\label{eq:qe}
\end{equation}
QE reranking scores each candidate independently given the source sentence and therefore requires $|\mathcal{C}|$ evaluations of QE-model. 

Nevertheless, both QE reranking and MBR decoding methods conventionally generate the full candidate set before selecting a translation. 
They therefore use the full candidate budget even when a smaller candidate set would suffice for output selection..
\textsc{Quit} addresses this limitation by adaptively determining when to stop expanding the candidate set for each source sentence.

\section{Proposed Method: \textsc{Quit}}
\label{sec:proposed_methods}

Conventional reranking uses the same fixed candidate budget for every source sentence, even when additional candidates cease to improve translation quality.
We propose \textsc{Quit} to address this inefficiency.
Its goal is to stop when the quality of the selected output is close to that of the full-budget reranker's output.
Since this difference is actually unavailable during incremental generation, \textsc{Quit} uses recent reranking score stability as an uncertainty signal for deciding when to stop.

Formally, let $N_{\max}$ denote the maximum candidate budget.
Candidates are generated in batches of $k$, so after generation step $i$, a total of $ik$ candidates have been generated.
Assuming that $k$ divides $N_{\max}$, the candidate count grows as follows:
\begin{equation}
0 < k < 2k < \cdots < N_{\max}.
\end{equation}
For $i\in\{1,\ldots,N_{\max}/k\}$, define the partial candidate set after step $i$ as:
\begin{equation}
\mathcal{C}_i=\{h_1,\ldots,h_{ik}\},
\qquad |\mathcal{C}_i|=ik.
\end{equation}
For MBR, 
we use the current candidate set as the support set, i.e., $\mathcal{S}_i=\mathcal{C}_i$, so both sets expand at each generation step.
Let $r$ be any reranking function for which higher scores are better and $\hat{y}_i$ denote the translation selected after generation step $i$:
\begin{equation}
\hat{y}_i
= \mathop{\mathrm{argmax}}\limits_{h\in\mathcal{C}_i}
r(h;s,\mathcal{C}_i).
\label{eq:partial_reranking_output}
\end{equation}
We denote the best reranking score after step $i$:
\begin{equation}
R_i(s) = \max_{h\in\mathcal{C}_i} r(h;s,\mathcal{C}_i).
\label{eq:best_reranking_score}
\end{equation}

\paragraph{Early-stopping Risk}
Let $q(h,y_s^{\mathrm{ref}})$ evaluate candidate $h$ against a human reference translation $y_s^{\mathrm{ref}}$, with higher scores indicating better translation quality.
We define the early-stopping risk at generation step $i$ as:
\begin{equation}
\mathcal{L}_i(s) = \big|q(\hat y_{i},y_s^{\mathrm{ref}}) -q(\hat y_{N_{\max}/k},y_s^{\mathrm{ref}})\big|.
\label{eq:early_stopping_risk}
\end{equation}
This nonnegative quantity measures the absolute quality difference between two outputs selected by the same reranker; a small value means that stopping preserves the full-budget output's quality.
However, $\mathcal{L}_i(s)$ cannot be computed during inference because both the human reference and the full-budget output are unavailable.

\begin{algorithm}[t]
\caption{\textsc{Quit}}
\label{alg:quit}
\DontPrintSemicolon
\Procedure{\textsc{Quit}$(s,r,k,w,\alpha,N_{\max})$}{
    \Comment*[l]{Initialize the candidate set and score buffer}
    $\mathcal{C} \gets \varnothing$\;
    $\mathcal{R} \gets [\,]$\;
    $i \gets 0$\;
    \While{$ik < N_{\max}$}{
        $i \gets i+1$\;
        Generate $k$ new candidates and add them to $\mathcal{C}$\;
        $\hat{y} \gets \mathop{\mathrm{argmax}}\limits_{h\in\mathcal{C}}
        r(h;s,\mathcal{C})$\;
        $R \gets \max\limits_{h\in\mathcal{C}}r(h;s,\mathcal{C})$\;
        Append $R$ to $\mathcal{R}$\;
        \If{$|\mathcal{R}|=w$}{
            $\Delta \gets \max \mathcal{R}-\min \mathcal{R}$\;
            \eIf{$\Delta \leq\alpha$}{
                \Return $\hat{y}$\;
            }{
                \Comment*[l]{Start a new non-overlapping window}
                $\mathcal{R} \gets [\,]$\;
            }
        }
    }
    \Return $\hat{y}$\;
}
\end{algorithm}
\paragraph{Reranking-based Risk Proxy}
To approximate the early-stopping risk, we use the reranker's own score.
We define the corresponding risk proxy as:
\begin{equation}
\begin{aligned}
\mathcal{L}_i^{\mathrm{proxy}}(s)
&=
\big|
r(\hat y_i;s,\mathcal{C}_i)
-
r(\hat y_{N_{\max}/k};s,\mathcal{C}_{N_{\max}/k})
\big| \\
&=
\big|
R_i(s)-R_{N_{\max}/k}(s)
\big|.
\end{aligned}
\label{eq:proxy_early_stopping_risk}
\end{equation}
This replaces the reference-based quality difference with the corresponding reranking-score difference.
During incremental inference, $\mathcal{L}_i^{\mathrm{proxy}}(s)$ remains unavailable because $R_{N_{\max}/k}(s)$ is only observed at the full candidate budget.

\paragraph{Within-window Uncertainty}
To obtain an observable signal during inference, \textsc{Quit} uses the recent variation of the reranking-score trajectory $\{R_i(s)\}$.
We group the generation steps into non-overlapping windows of size $w$.
The $b$-th window contains the step indices
$\mathcal{B}_b = \left\{ i: (b-1)w < i \le bw \right\}$,
where 
$b\in\{1,\ldots,\lfloor N_{\max}/(kw)\rfloor\}$.
We measure its within-window variation by the range:
\begin{equation}
\Delta_b^{R}(s)
=
\max_{j\in\mathcal{B}_b}R_j(s)
-
\min_{j\in\mathcal{B}_b}R_j(s).
\label{eq:reranking_score_range}
\end{equation}
A large $\Delta_b^{R}(s)$ indicates that the best reranking score has improved substantially within the recent window, whereas a small value indicates local stability, suggesting that further sampling may have limited benefit.
Accordingly, $\Delta_b^{R}(s)$ serves as a local uncertainty signal that provides a partial indication of the early-stopping risk at the end of window $b$: $\mathcal{L}_{bw}(s)$.

Given a convergence threshold $\alpha$, \textsc{Quit} stops at
\begin{equation}
T
=
\min
\left\{
bw:\Delta_b^{R}(s)\leq\alpha
\right\},
\label{eq:stopping_time}
\end{equation}
and returns $\hat y_T$.
If the condition is never satisfied, generation continues until the full candidate budget is reached.
Algorithm~\ref{alg:quit} summarizes the procedure.

\FloatBarrier

\begin{table*}[t]
\centering
\footnotesize
\setlength{\tabcolsep}{3.5pt}
\begin{tabular}{@{}lllcccccc@{}}
\toprule
\textbf{NMT Model} & \textbf{Reranker} & \textbf{Acceleration} & \textbf{Speedup $\uparrow$} & \textbf{Rerank. $\uparrow$} & \textbf{ChrF++ $\uparrow$} & \textbf{xCOMET $\uparrow$} & \textbf{MetricX $\downarrow$} & \textbf{GEMBA $\uparrow$} \\
\midrule
\multirow{10}{*}{Qwen3} & \multirow{6}{*}{MBR} & Unaccelerated & 1.00 & .8357 & 42.09 & .7520 & 5.17 & -7.82 \\
 &  & PruneMBR & 1.05 & ~~~.8355\textsuperscript{\dag\dag} & ~~~42.07\textsuperscript{\dag\dag} & ~~~.7519\textsuperscript{\dag\dag} & ~~~5.17\textsuperscript{\dag\dag} & ~~~-7.81\textsuperscript{\dag\dag} \\
 &  & PMBR & 1.03 & ~~~.8352\textsuperscript{\dag\dag} & ~~~42.11\textsuperscript{\dag\dag} & ~~~.7516\textsuperscript{\dag\dag} & ~~~5.17\textsuperscript{\dag\dag} & ~~~-7.81\textsuperscript{\dag\dag} \\
 &  & \textsc{Quit} ($w=4$) & 3.44 & ~.8315\textsuperscript{\dag} & ~~~41.94\textsuperscript{\dag\dag} & ~.7462\textsuperscript{\dag} & ~5.28\textsuperscript{\dag} & ~-8.03\textsuperscript{\dag} \\
 &  & \textsc{Quit} ($w=8$) & 1.74 & ~~~.8344\textsuperscript{\dag\dag} & ~~~42.04\textsuperscript{\dag\dag} & ~~~.7497\textsuperscript{\dag\dag} & ~~~5.19\textsuperscript{\dag\dag} & ~~~-7.87\textsuperscript{\dag\dag} \\
 &  & \textsc{Quit} ($w=16$) & 1.16 & ~~~.8355\textsuperscript{\dag\dag} & ~~~42.06\textsuperscript{\dag\dag} & ~~~.7518\textsuperscript{\dag\dag} & ~~~5.17\textsuperscript{\dag\dag} & ~~~-7.82\textsuperscript{\dag\dag} \\
\cmidrule{2-9}
 & \multirow{4}{*}{QE} & Unaccelerated & 1.00 & .8223 & 39.61 & .7585 & 4.96 & -7.91 \\
 &  & \textsc{Quit} ($w=4$) & 8.16 & .8067 & 40.72 & .7464 & 5.25 & ~-8.10\textsuperscript{\dag} \\
 &  & \textsc{Quit} ($w=8$) & 3.64 & .8133 & ~40.46\textsuperscript{\dag} & ~.7524\textsuperscript{\dag} & ~5.13\textsuperscript{\dag} & ~~~-7.92\textsuperscript{\dag\dag} \\
 &  & \textsc{Quit} ($w=16$) & 1.73 & ~.8185\textsuperscript{\dag} & ~40.21\textsuperscript{\dag} & ~~~.7572\textsuperscript{\dag\dag} & ~~~5.01\textsuperscript{\dag\dag} & ~~~-7.84\textsuperscript{\dag\dag} \\
\midrule
\multirow{10}{*}{TranslateGemma} & \multirow{6}{*}{MBR} & Unaccelerated & 1.00 & .8570 & 44.90 & .8521 & 3.15 & -3.14 \\
 &  & PruneMBR & 1.04 & ~~~.8570\textsuperscript{\dag\dag} & ~~~44.88\textsuperscript{\dag\dag} & ~~~.8520\textsuperscript{\dag\dag} & ~~~3.15\textsuperscript{\dag\dag} & ~~~-3.13\textsuperscript{\dag\dag} \\
 &  & PMBR & 1.03 & ~~~.8566\textsuperscript{\dag\dag} & ~~~44.91\textsuperscript{\dag\dag} & ~~~.8518\textsuperscript{\dag\dag} & ~~~3.15\textsuperscript{\dag\dag} & ~~~-3.16\textsuperscript{\dag\dag} \\
 &  & \textsc{Quit} ($w=4$) & 5.23 & ~~~.8557\textsuperscript{\dag\dag} & ~~~44.88\textsuperscript{\dag\dag} & ~~~.8502\textsuperscript{\dag\dag} & ~~~3.18\textsuperscript{\dag\dag} & ~~~-3.16\textsuperscript{\dag\dag} \\
 &  & \textsc{Quit} ($w=8$) & 2.66 & ~~~.8566\textsuperscript{\dag\dag} & ~~~44.88\textsuperscript{\dag\dag} & ~~~.8509\textsuperscript{\dag\dag} & ~~~3.16\textsuperscript{\dag\dag} & ~~~-3.16\textsuperscript{\dag\dag} \\
 &  & \textsc{Quit} ($w=16$) & 1.51 & ~~~.8569\textsuperscript{\dag\dag} & ~~~44.87\textsuperscript{\dag\dag} & ~~~.8513\textsuperscript{\dag\dag} & ~~~3.15\textsuperscript{\dag\dag} & ~~~-3.16\textsuperscript{\dag\dag} \\
\cmidrule{2-9}
 & \multirow{4}{*}{QE} & Unaccelerated & 1.00 & .8376 & 43.59 & .8570 & 3.14 & -3.12 \\
 &  & \textsc{Quit} ($w=4$) & 8.92 & .8295 & 44.57 & ~.8534\textsuperscript{\dag} & ~3.17\textsuperscript{\dag} & ~-2.98\textsuperscript{\dag} \\
 &  & \textsc{Quit} ($w=8$) & 4.01 & .8327 & ~44.37\textsuperscript{\dag} & ~~~.8553\textsuperscript{\dag\dag} & ~~~3.15\textsuperscript{\dag\dag} & ~-3.01\textsuperscript{\dag} \\
 &  & \textsc{Quit} ($w=16$) & 1.88 & ~.8355\textsuperscript{\dag} & ~44.24\textsuperscript{\dag} & ~~~.8570\textsuperscript{\dag\dag} & ~~~3.14\textsuperscript{\dag\dag} & ~-3.00\textsuperscript{\dag} \\
\midrule
\multirow{10}{*}{Hy-MT2} & \multirow{6}{*}{MBR} & Unaccelerated & 1.00 & .8731 & 49.86 & .8867 & 2.92 & -1.79 \\
 &  & PruneMBR & 1.02 & ~~~.8732\textsuperscript{\dag\dag} & ~~~49.87\textsuperscript{\dag\dag} & ~~~.8869\textsuperscript{\dag\dag} & ~~~2.92\textsuperscript{\dag\dag} & ~~~-1.80\textsuperscript{\dag\dag} \\
 &  & PMBR & 1.02 & ~~~.8728\textsuperscript{\dag\dag} & ~~~49.87\textsuperscript{\dag\dag} & ~~~.8866\textsuperscript{\dag\dag} & ~~~2.93\textsuperscript{\dag\dag} & ~~~-1.79\textsuperscript{\dag\dag} \\
 &  & \textsc{Quit} ($w=4$) & 5.42 & ~~~.8723\textsuperscript{\dag\dag} & ~~~49.85\textsuperscript{\dag\dag} & ~~~.8863\textsuperscript{\dag\dag} & ~~~2.94\textsuperscript{\dag\dag} & ~~~-1.78\textsuperscript{\dag\dag} \\
 &  & \textsc{Quit} ($w=8$) & 2.66 & ~~~.8728\textsuperscript{\dag\dag} & ~~~49.86\textsuperscript{\dag\dag} & ~~~.8867\textsuperscript{\dag\dag} & ~~~2.93\textsuperscript{\dag\dag} & ~~~-1.79\textsuperscript{\dag\dag} \\
 &  & \textsc{Quit} ($w=16$) & 1.50 & ~~~.8731\textsuperscript{\dag\dag} & ~~~49.83\textsuperscript{\dag\dag} & ~~~.8866\textsuperscript{\dag\dag} & ~~~2.93\textsuperscript{\dag\dag} & ~~~-1.80\textsuperscript{\dag\dag} \\
\cmidrule{2-9}
 & \multirow{4}{*}{QE} & Unaccelerated & 1.00 & .8366 & 47.67 & .8801 & 2.98 & -2.37 \\
 &  & \textsc{Quit} ($w=4$) & 9.17 & .8290 & 48.77 & ~~~.8814\textsuperscript{\dag\dag} & ~~~2.98\textsuperscript{\dag\dag} & -2.03 \\
 &  & \textsc{Quit} ($w=8$) & 4.12 & .8317 & ~48.54\textsuperscript{\dag} & ~~~.8821\textsuperscript{\dag\dag} & ~~~2.96\textsuperscript{\dag\dag} & -2.06 \\
 &  & \textsc{Quit} ($w=16$) & 1.94 & ~.8342\textsuperscript{\dag} & ~48.19\textsuperscript{\dag} & ~~~.8816\textsuperscript{\dag\dag} & ~~~2.96\textsuperscript{\dag\dag} & ~-2.16\textsuperscript{\dag} \\
\bottomrule
\end{tabular}
\caption{Main results on WMT24 with a maximum candidate budget $N_{\max} = 512$. The reranking score (Rerank.) was computed using COMET-22 for MBR decoding and CometKiwi-22 for QE reranking. \textsuperscript{\dag} marks scores that are statistically equivalent to the unaccelerated baseline within a prespecified margin of $0.05\sigma$, and \textsuperscript{\dag\dag} marks equivalence within a stricter margin of $0.02\sigma$. Speedup was measured end-to-end relative to the baseline.}\label{tab:main_results_wmt24}
\end{table*}

\begin{table*}[t]
\centering
\footnotesize
\setlength{\tabcolsep}{3.5pt}
\begin{tabular}{@{}lllcccccc@{}}
\toprule
\textbf{NMT Model} & \textbf{Reranker} & \textbf{Acceleration} & \textbf{Speedup $\uparrow$} & \textbf{Rerank. $\uparrow$} & \textbf{ChrF++ $\uparrow$} & \textbf{xCOMET $\uparrow$} & \textbf{MetricX $\downarrow$} & \textbf{GEMBA $\uparrow$} \\
\midrule
\multirow{10}{*}{Qwen3} & \multirow{6}{*}{MBR} & Unaccelerated & 1.00 & .7345 & 34.06 & .3696 & 10.86 & -18.11 \\
 &  & PruneMBR & 1.02 & ~~~.7344\textsuperscript{\dag\dag} & ~~~34.04\textsuperscript{\dag\dag} & ~~~.3692\textsuperscript{\dag\dag} & ~~~10.86\textsuperscript{\dag\dag} & ~~~-18.10\textsuperscript{\dag\dag} \\
 &  & PMBR & 1.01 & ~~~.7339\textsuperscript{\dag\dag} & ~~~34.07\textsuperscript{\dag\dag} & ~~~.3689\textsuperscript{\dag\dag} & ~~~10.88\textsuperscript{\dag\dag} & ~~~-18.16\textsuperscript{\dag\dag} \\
 &  & \textsc{Quit} ($w=4$) & 2.81 & ~.7280\textsuperscript{\dag} & ~~~33.89\textsuperscript{\dag\dag} & ~.3621\textsuperscript{\dag} & ~11.07\textsuperscript{\dag} & ~-18.36\textsuperscript{\dag} \\
 &  & \textsc{Quit} ($w=8$) & 1.47 & ~~~.7322\textsuperscript{\dag\dag} & ~~~33.98\textsuperscript{\dag\dag} & ~~~.3671\textsuperscript{\dag\dag} & ~~~10.95\textsuperscript{\dag\dag} & ~~~-18.19\textsuperscript{\dag\dag} \\
 &  & \textsc{Quit} ($w=16$) & 1.10 & ~~~.7340\textsuperscript{\dag\dag} & ~~~34.04\textsuperscript{\dag\dag} & ~~~.3689\textsuperscript{\dag\dag} & ~~~10.88\textsuperscript{\dag\dag} & ~~~-18.17\textsuperscript{\dag\dag} \\
\cmidrule{2-9}
 & \multirow{4}{*}{QE} & Unaccelerated & 1.00 & .7106 & 30.32 & .3705 & 10.10 & -18.24 \\
 &  & \textsc{Quit} ($w=4$) & 7.75 & .6818 & 31.81 & .3532 & 10.88 & -18.60 \\
 &  & \textsc{Quit} ($w=8$) & 3.43 & .6935 & 31.28 & ~.3613\textsuperscript{\dag} & 10.54 & ~-18.37\textsuperscript{\dag} \\
 &  & \textsc{Quit} ($w=16$) & 1.64 & .7036 & ~30.80\textsuperscript{\dag} & ~.3665\textsuperscript{\dag} & ~10.26\textsuperscript{\dag} & ~~~-18.22\textsuperscript{\dag\dag} \\
\midrule
\multirow{10}{*}{TranslateGemma} & \multirow{6}{*}{MBR} & Unaccelerated & 1.00 & .8129 & 36.25 & .5731 & 6.06 & -6.74 \\
 &  & PruneMBR & 1.01 & ~~~.8129\textsuperscript{\dag\dag} & ~~~36.25\textsuperscript{\dag\dag} & ~~~.5735\textsuperscript{\dag\dag} & ~~~6.06\textsuperscript{\dag\dag} & ~~~-6.71\textsuperscript{\dag\dag} \\
 &  & PMBR & 1.01 & ~~~.8125\textsuperscript{\dag\dag} & ~~~36.22\textsuperscript{\dag\dag} & ~~~.5717\textsuperscript{\dag\dag} & ~~~6.08\textsuperscript{\dag\dag} & ~-6.85\textsuperscript{\dag} \\
 &  & \textsc{Quit} ($w=4$) & 4.76 & ~.8099\textsuperscript{\dag} & ~~~36.13\textsuperscript{\dag\dag} & ~.5673\textsuperscript{\dag} & ~6.17\textsuperscript{\dag} & ~-6.94\textsuperscript{\dag} \\
 &  & \textsc{Quit} ($w=8$) & 2.61 & ~~~.8112\textsuperscript{\dag\dag} & ~~~36.18\textsuperscript{\dag\dag} & ~.5690\textsuperscript{\dag} & ~6.12\textsuperscript{\dag} & ~-6.87\textsuperscript{\dag} \\
 &  & \textsc{Quit} ($w=16$) & 1.51 & ~~~.8123\textsuperscript{\dag\dag} & ~~~36.19\textsuperscript{\dag\dag} & ~~~.5713\textsuperscript{\dag\dag} & ~~~6.10\textsuperscript{\dag\dag} & ~~~-6.79\textsuperscript{\dag\dag} \\
\cmidrule{2-9}
 & \multirow{4}{*}{QE} & Unaccelerated & 1.00 & .7721 & 36.17 & .5670 & 6.11 & -6.80 \\
 &  & \textsc{Quit} ($w=4$) & 8.09 & .7608 & ~~~36.15\textsuperscript{\dag\dag} & ~.5601\textsuperscript{\dag} & ~6.23\textsuperscript{\dag} & ~-7.03\textsuperscript{\dag} \\
 &  & \textsc{Quit} ($w=8$) & 3.55 & .7657 & ~~~36.16\textsuperscript{\dag\dag} & ~~~.5645\textsuperscript{\dag\dag} & ~6.18\textsuperscript{\dag} & ~-6.86\textsuperscript{\dag} \\
 &  & \textsc{Quit} ($w=16$) & 1.69 & ~.7694\textsuperscript{\dag} & ~~~36.20\textsuperscript{\dag\dag} & ~~~.5654\textsuperscript{\dag\dag} & ~~~6.14\textsuperscript{\dag\dag} & ~~~-6.81\textsuperscript{\dag\dag} \\
\midrule
\multirow{10}{*}{Hy-MT2} & \multirow{6}{*}{MBR} & Unaccelerated & 1.00 & .8298 & 41.19 & .6281 & 5.62 & -4.25 \\
 &  & PruneMBR & 1.01 & ~~~.8298\textsuperscript{\dag\dag} & ~~~41.20\textsuperscript{\dag\dag} & ~~~.6278\textsuperscript{\dag\dag} & ~~~5.62\textsuperscript{\dag\dag} & ~~~-4.19\textsuperscript{\dag\dag} \\
 &  & PMBR & 1.01 & ~~~.8294\textsuperscript{\dag\dag} & ~~~41.19\textsuperscript{\dag\dag} & ~~~.6279\textsuperscript{\dag\dag} & ~~~5.63\textsuperscript{\dag\dag} & ~~~-4.23\textsuperscript{\dag\dag} \\
 &  & \textsc{Quit} ($w=4$) & 4.46 & ~~~.8279\textsuperscript{\dag\dag} & ~~~41.19\textsuperscript{\dag\dag} & ~.6226\textsuperscript{\dag} & ~5.69\textsuperscript{\dag} & ~-4.38\textsuperscript{\dag} \\
 &  & \textsc{Quit} ($w=8$) & 2.36 & ~~~.8289\textsuperscript{\dag\dag} & ~~~41.18\textsuperscript{\dag\dag} & ~~~.6262\textsuperscript{\dag\dag} & ~~~5.65\textsuperscript{\dag\dag} & ~~~-4.25\textsuperscript{\dag\dag} \\
 &  & \textsc{Quit} ($w=16$) & 1.40 & ~~~.8295\textsuperscript{\dag\dag} & ~~~41.17\textsuperscript{\dag\dag} & ~~~.6279\textsuperscript{\dag\dag} & ~~~5.63\textsuperscript{\dag\dag} & ~~~-4.22\textsuperscript{\dag\dag} \\
\cmidrule{2-9}
 & \multirow{4}{*}{QE} & Unaccelerated & 1.00 & .7786 & 40.78 & .6219 & 5.71 & -4.88 \\
 &  & \textsc{Quit} ($w=4$) & 8.12 & .7678 & ~~~40.89\textsuperscript{\dag\dag} & ~.6151\textsuperscript{\dag} & ~5.81\textsuperscript{\dag} & ~-4.65\textsuperscript{\dag} \\
 &  & \textsc{Quit} ($w=8$) & 3.59 & .7725 & ~~~40.90\textsuperscript{\dag\dag} & ~.6184\textsuperscript{\dag} & ~5.77\textsuperscript{\dag} & ~-4.74\textsuperscript{\dag} \\
 &  & \textsc{Quit} ($w=16$) & 1.70 & ~.7761\textsuperscript{\dag} & ~~~40.84\textsuperscript{\dag\dag} & ~~~.6210\textsuperscript{\dag\dag} & ~~~5.74\textsuperscript{\dag\dag} & ~~~-4.83\textsuperscript{\dag\dag} \\
\bottomrule
\end{tabular}
\caption{Main results on WMT25, following the conventions in Table~\ref{tab:main_results_wmt24}.}\label{tab:main_results_wmt25}
\end{table*}

\section{Experiments}
\label{sec:exp}
\subsection{Experimental Setup}
\label{sec:exp_setup}

\paragraph{Datasets and NMT Models}
We used the complete test sets from the WMT24 \cite{wmt24} and WMT25 \cite{wmt25} General Translation Shared Tasks, which together cover 19 language pairs.
We evaluated three NMT models across two categories.
Qwen3 (\code{Qwen3-8B}) \cite{qwen3} is a general-purpose LLM without translation-specific fine-tuning, whereas TranslateGemma (\code{translategemma-12b-it}) \cite{translategemma} and Hy-MT2 (\code{Hy-MT2-30B-A3B}) \cite{hymt2} are translation-specialized models.
For each source sentence, we set the maximum candidate budget to $N_{\max}=512$ and used ancestral sampling with a temperature of 1.2.

\paragraph{Acceleration Baselines}
For MBR, we compared our method against PruneMBR \cite{CPMBR} and PMBR \cite{pmbr}, which reduce MBR decoding computation while retaining the full candidate budget.
For PMBR, we set the ALS rank to $r=10$ and the regularization coefficient to $\lambda=0.1$, with a computation budget of $1/8$.
For PruneMBR, we used a confidence threshold of $\alpha=0.99$, $n=500$ bootstrap samples, and a pseudo-reference sample-size schedule of $[8,16,32,64,128,256,512]$.

\paragraph{\textsc{Quit} Hyperparameters}
We fixed the generation batch size to $k=8$ and the convergence threshold to $\alpha=10^{-3}$, and evaluated window sizes $w\in\{2,4,6,8,10,12,14,16\}$.
The main results report $w\in\{4,8,16\}$.\footnote{Additional analyses of $k$ and $\alpha$ are provided in Appendices~\ref{sec:batch_size_analyze} and~\ref{sec:alpha_ablation}.
The complete trends across window sizes are reported in Appendix~\ref{sec:complete_window_sweep}.
}
%An ablation study of $\alpha$ is reported in Appendix~\ref{sec:alpha_ablation}.

\paragraph{Rerankers}
We considered two widely used reranking paradigms in NMT: MBR decoding and QE reranking.
For MBR, we used COMET-22 (\code{wmt22-comet-da}) \cite{comet22} as the pairwise utility function, and additionally report results using ChrF \cite{chrf} as an alternative utility in Appendix \ref{sec:chrf_utility}.
For QE reranking, we used CometKiwi-22 (\code{wmt22-cometkiwi-da}) \cite{cometkiwi22} as the QE model.
More recent metric models that can be used for reranking, such as MetricX\cite{metricx24}, are substantially larger and would be prohibitively expensive to evaluate at our experimental scale; we therefore excluded them.

\paragraph{Quality Evaluation}
We evaluated translation quality using three complementary classes of automatic metrics that correlate strongly with human judgments.
We used ChrF++ \cite{chrfpp} as a surface-form metric; xCOMET (\code{XCOMET-XXL}) \cite{xcomet} and MetricX (\code{metricx-24-hybrid-xl-v2p6}) \cite{metricx24} as learned neural metrics; and GEMBA-MQM (hereafter GEMBA) \cite{gemba}, instantiated with Gemma-4-31B (\code{gemma-4-31B-it}) \cite{gemmat4}, as an LLM-based metric.
Because GEMBA is methodologically distinct from the COMET-family rerankers, it provides an independent check and reduces the risk that the evaluator favors outputs optimized by a similar metric.
We also report the COMET-22 and CometKiwi-22 reranking scores to measure how closely early stopping preserves each reranker's preferences.
To assess whether acceleration affects translation quality, we ran paired-bootstrap equivalence tests using two one-sided tests (TOST) \cite{tost} with $10{,}000$ resamples against the unaccelerated baseline for every metric.
For each metric, we defined $\sigma$ as the baseline segment-level standard deviation and used equivalence margins of $0.05\sigma$ and $0.02\sigma$, with $0.02\sigma$ serving as the stricter margin.

\paragraph{Efficiency Evaluation}
For efficiency, we report end-to-end speedup relative to the unaccelerated baseline.
End-to-end time includes both candidate generation and reranking, with all measurements conducted under the same hardware configuration using two NVIDIA RTX A6000 GPUs.

\subsection{Experimental Results}\label{sec:results}

Tables~\ref{tab:main_results_wmt24} and~\ref{tab:main_results_wmt25} summarize the main experimental results on the two test sets, reporting both translation quality and end-to-end efficiency.
Among the window sizes shown in the main tables, $w=8$ provided a good balance between quality and efficiency, yielding speedups of $1.47$--$2.66\times$ for MBR decoding and $3.43$--$4.12\times$ for QE reranking across the two datasets and three NMT models.
For most quality metrics, the resulting scores were statistically equivalent to those of the unaccelerated baseline, with several scores improving over the baseline.
Some QE reranking scores did not pass the equivalence test.
For QE, each candidate's score is independent of the other candidates, so the best reranking score is nondecreasing with candidate-set size and inherently favors the full 512-candidate set.
Increasing the window size to 16 generally reduced speedup but brought reranking scores and external quality scores closer to those of the unaccelerated baseline.
%The complete trends across window sizes are reported in Appendix~\ref{sec:complete_window_sweep}.
In contrast, PruneMBR and PMBR achieved an end-to-end speedup of at most $1.05\times$ because candidate generation remains unchanged.

\section{Discussion}
\label{sec:discussion}

\subsection{Does the Uncertainty Score Identify Early-stopping Risk?}
\label{sec:prr_risk}
We examine whether \textsc{Quit}'s uncertainty score $\Delta_b^{R}(s)$ ranks the early-stopping risk $\mathcal{L}_{bw}(s)$, defined in Equation~\ref{eq:early_stopping_risk}.
Following the prediction-rejection ratio (PRR) framework \cite{uq}, we treated each source--window pair as an instance, using $\mathcal{L}_{bw}(s)$ as the risk and $\Delta_b^{R}(s)$ as the uncertainty score.
Rejecting instances in descending order of uncertainty gives a curve of the mean early-stopping risk among retained instances.
We computed PRR as follows:
\begin{equation}
\mathrm{PRR}
=\frac{\mathrm{AUC}_{\mathrm{rnd}}-\mathrm{AUC}_{\mathrm{uns}}}
{\mathrm{AUC}_{\mathrm{rnd}}-\mathrm{AUC}_{\mathrm{oracle}}},
\label{eq:prr}
\end{equation}
where $\mathrm{AUC}_{\mathrm{uns}}$ is the area under the curve, $\mathrm{AUC}_{\mathrm{oracle}}$ uses the ranking by $\mathcal{L}_{bw}(s)$ itself, and $\mathrm{AUC}_{\mathrm{rnd}}$ uses random ranking.
A value of 1 indicates oracle-equivalent early-stopping-risk ranking, whereas 0 indicates random ranking.

\begin{table*}[t]
\centering
\small
\setlength{\tabcolsep}{5pt}
\begin{tabular}{@{}llcccc@{}}
\toprule
\multirow{2}{*}{\textbf{Test set}} &
\multirow{2}{*}{\textbf{NMT model}} &
\multicolumn{2}{c}{\textbf{MBR}} &
\multicolumn{2}{c}{\textbf{QE}} \\
\cmidrule(lr){3-4}\cmidrule(lr){5-6}
& & \textbf{xCOMET} & \textbf{GEMBA} & \textbf{xCOMET} & \textbf{GEMBA} \\
\midrule
WMT24 & Qwen3 & 0.412 & 0.397 & 0.174 & 0.152 \\
WMT24 & TranslateGemma & 0.384 & 0.390 & 0.191 & 0.176 \\
WMT24 & Hy-MT2 & 0.493 & 0.560 & 0.176 & 0.153 \\
WMT25 & Qwen3 & 0.252 & 0.101 & 0.172 & 0.134 \\
WMT25 & TranslateGemma & 0.269 & 0.322 & 0.152 & 0.142 \\
WMT25 & Hy-MT2 & 0.307 & 0.479 & 0.168 & 0.175 \\
\midrule
\multicolumn{2}{l}{\textbf{Average}} & \textbf{0.353} & \textbf{0.375} & \textbf{0.172} & \textbf{0.155} \\
\bottomrule
\end{tabular}
\caption{Pooled PRR with uncertainty $\Delta_b^{R}(s)$ and early-stopping risk $\mathcal{L}_{bw}(s)$ (Equation~\ref{eq:early_stopping_risk}), $w=8$. Higher PRR is better.}
\label{tab:prr_w8}
\end{table*}
Table~\ref{tab:prr_w8} reports pooled PRR for $w=8$ across WMT24 and WMT25, and all three NMT models.
PRR is positive in all 24 settings.
The signal is more informative on average for MBR than for QE, but remains well below oracle ranking for both.
These results support $\Delta_b^{R}(s)$ as an informative, imperfect indicator of early-stopping risk.
\footnote{Appendix~\ref{sec:prr_computation} gives the evaluation procedure and corresponding results using $\mathcal{L}^{proxy}_{bw}(s)$ as the early-stopping risk.}

\subsection{Where does \textsc{Quit} Stop?}
\label{sec:stopping_distribution}

\begin{figure*}[t]
    \centering
    \includegraphics[width=.85\textwidth]{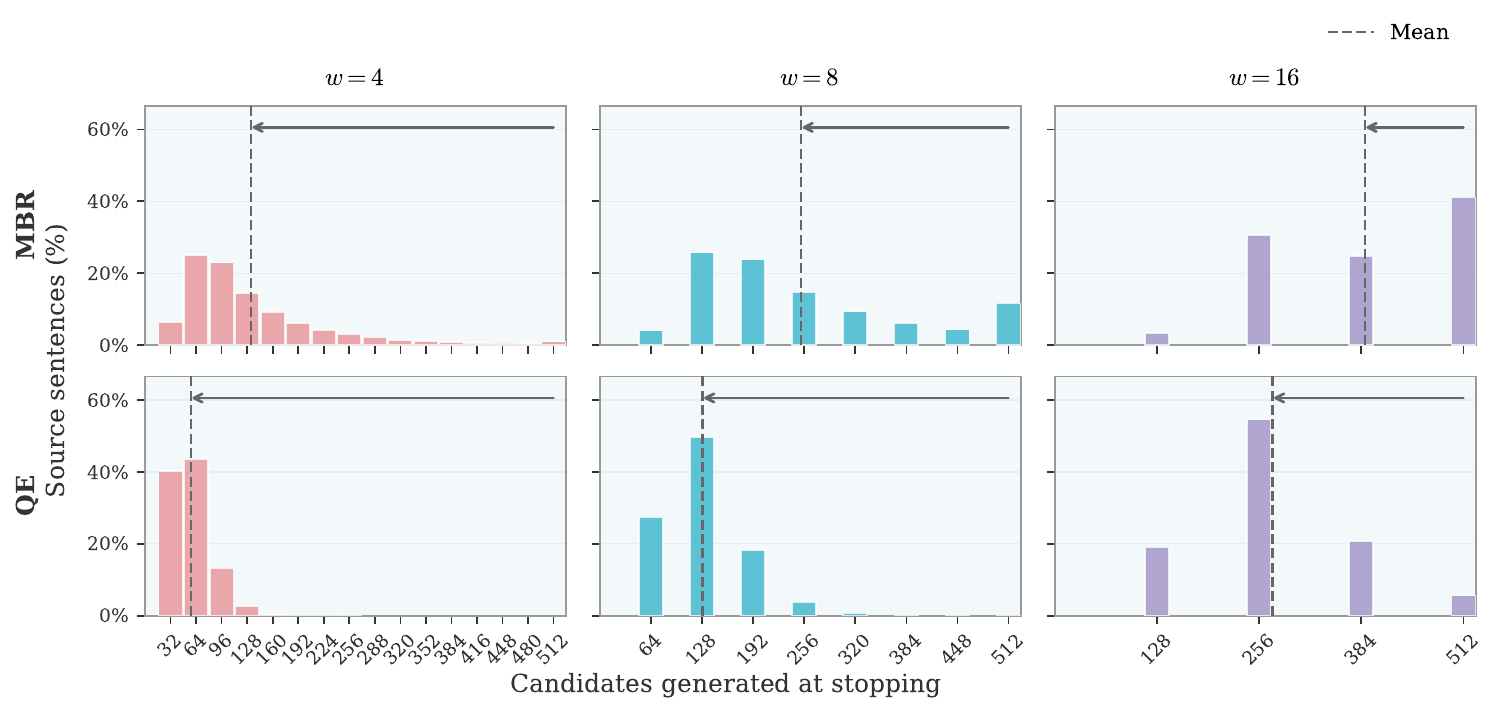}
    \caption{Distribution of candidate-set sizes when \textsc{Quit} stops for MBR decoding and QE reranking. Bars show the percentage of source sentences for each window size. The bar at 512 includes sentences for which generation reaches the full candidate budget.
    Bars are placed at the possible stopping points, i.e., multiples of $kw$ candidates ($kw=32$, $64$, and $128$ for $w=4$, $8$, and $16$).}
    \label{fig:stopping_distribution}
\end{figure*}

Figure~\ref{fig:stopping_distribution} shows that QE generally stops with relatively small candidate sets, whereas MBR stopping positions are more broadly distributed across the candidate budget.
This difference is consistent with how the two reranking scores evolve
\footnote{Appendix~\ref{sec:score_evolution} provides further analysis.}
: QE scores for existing candidates are unaffected by later candidates, whereas expanding MBR's support set changes existing candidates' scores and may delay stabilization of the best score. 
Larger windows generally delay stopping, particularly for MBR, producing behavior closer to full-budget reranking at the cost of smaller speedups.

\subsection{Where do the Efficiency Gains Come from?}
\label{sec:runtime_breakdown}

\begin{table}[t]
\centering
\small
\setlength{\tabcolsep}{4pt}
\begin{tabular}{@{}lcrrr@{}}
\toprule
\multirow{2}{*}{\textbf{Method}} &
\multirow{2}{*}{$\boldsymbol{w}$} &
\multicolumn{3}{c}{\textbf{Time per segment (s) $\downarrow$}} \\
\cmidrule(lr){3-5}
& & \textbf{Gen.} & \textbf{Rerank.} & \textbf{Total} \\
\midrule
Unaccelerated MBR & $\infty$ & 53.52 & 1.40 & 54.92 \\
PMBR           & --       & 53.52 & 0.53 & 54.05 \\
PruneMBR       & --       & 53.52 & 0.33 & 53.85 \\
\textsc{Quit}  & 4        & 12.95 & 0.14 & 13.08 \\
\textsc{Quit}  & 8        & 24.69 & 0.43 & 25.13 \\
\textsc{Quit}  & 16       & 39.58 & 0.88 & 40.46 \\
\bottomrule
\end{tabular}
\caption{Average component-wise MBR runtime per source segment. Times were pooled over WMT24 and WMT25, and the three NMT models, weighting each test set by its number of source segments. Complete aggregate wall-clock times are given in Appendix~\ref{sec:runtime_details}.}
\label{tab:runtime_breakdown}
\end{table}

Table~\ref{tab:runtime_breakdown} separates the cost of candidate generation from that of MBR reranking.
For the unaccelerated MBR baseline, candidate generation requires 53.52 seconds per segment on average, compared with only 1.40 seconds for reranking.
The results reveal that candidate generation, rather than reranking, is the dominant bottleneck, accounting for nearly all end-to-end latency in unaccelerated MBR.
Consequently, PMBR and PruneMBR substantially reduce the cost of reranking itself but yield little end-to-end improvement because they still generate the full candidate set.
This highlights a broader limitation of reranking-only acceleration: as candidate generation becomes more expensive, further reducing the comparatively small reranking cost offers increasingly limited end-to-end benefits.

\section{Related Work}
\label{sec:related_work}
\paragraph{Reranking for NMT}
Although selecting the most probable candidate is a natural decoding strategy, model probability is often poorly aligned with translation quality \cite{nmt_model_error,nmt_uncertainty,reranking_nerual_metric}, motivating the use of external criteria for output selection.
Against this background, MBR decoding, a decision rule widely studied in statistical machine translation, has been revisited for NMT.
Early work on neural MBR decoding typically used surface-form metrics \cite{is_map_need,nmt_mbr_surface2}, such as sentence-level BLEU \cite{bleu} or ChrF \cite{chrf,chrfpp}, as utility functions for comparing candidates with sampled support hypotheses.
As learned metrics have shown stronger agreement with human judgments, more recent work has increasingly adopted neural metrics as MBR utilities, allowing the decision rule to capture semantic similarity beyond lexical overlap \cite{reranking_nerual_metric,qa_reranking,smbr,kamigaito-etal-2025-diversity}.
In parallel, advances in reference-free QE have enabled QE reranking, which directly scores each source--hypothesis pair and selects the highest-scoring candidate without relying on support hypotheses \cite{qa_reranking}.

\paragraph{Efficient Methods for NMT Reranking}
Despite their effectiveness, both MBR decoding and QE reranking can be prohibitively expensive in latency-sensitive settings because they conventionally require a large, fully generated candidate set.
Existing efficiency methods have focused primarily on MBR decoding by reducing the number or cost of utility-function evaluations, for example, by pruning or approximating candidate--support comparisons \cite{pmbr,CentroidMBR,CPMBR,hfcpmbr}.
These methods optimize the reranking stage but generally retain the full candidate budget.
This limitation is increasingly consequential as LLM-based systems grow larger: candidate generation can dominate end-to-end inference time, even when reranking is performed naively.
Moreover, QE reranking avoids MBR's quadratic pairwise comparisons but still pays the generation and scoring costs for every candidate in the fixed budget.
To the best of our knowledge, \textsc{Quit} is the first acceleration method for NMT reranking that adaptively stops candidate generation and thereby reduces both candidate-generation and reranking costs.

\section{Conclusions and Future Work}
\label{sec:conclusions}
In this paper, we proposed \textsc{Quit}, an uncertainty-guided early-stopping method that jointly reduces candidate-generation and reranking costs.
It uses local reranking-score stability to decide when to stop, with the aim of keeping the selected output's quality close to the quality achieved by full-budget reranking.
Across three NMT models and 19 language pairs, \textsc{Quit} with $w=8$ achieved end-to-end speedups of $1.47$--$2.66\times$ for MBR decoding and $3.43$--$4.12\times$ for QE reranking while remaining within the tested equivalence margins for most automatic metric scores.

Future work will explore early-stopping signal that do not require manually specified hyperparameters and evaluate the effectiveness of \textsc{Quit} on tasks beyond machine translation.

\section*{Limitations}
First, although our experiments cover three NMT models and 19 language pairs from WMT24 and WMT25, our conclusions remain tied to these models and test distributions.
The observed speedups depend on the relative costs of candidate generation and reranking, which are influenced by model size and data characteristics.
If candidate generation becomes substantially cheaper in future NMT systems, reranking may instead dominate end-to-end latency, and the speedups achieved by \textsc{Quit} could differ considerably from those reported here.

Second, \textsc{Quit} assumes that the candidate set can be generated incrementally, with newly sampled candidates appended to the existing set.
This assumption does not hold directly for generation procedures such as beam search, in which hypotheses are jointly expanded and the resulting candidate sets are not necessarily nested.
Adapting \textsc{Quit} to such procedures requires further investigation.

Finally, using the uncertainty score in Equation~\ref{eq:reranking_score_range} assumes that reranking scores obtained from different candidate sets are comparable.
Although this assumption holds for the reranking methods considered in our experiments, future reranking objectives may produce scores whose scales vary with the candidate set.
In such cases, the current stopping criterion may require score normalization or calibration.
Even with comparable scores, local stability cannot guarantee low early-stopping risk: later candidates or changes in the reranker's selection may still alter output quality.

%\section*{Acknowledgments}

\bibliography{custom}

\clearpage

\appendix

\section{Effect of Batch Size $k$}
\label{sec:batch_size_analyze}
QUIT generates $k$ candidates at each step and evaluates score stability over a window of $w$ consecutive observations; therefore, each window spans $kw$ newly generated candidates. 

\begin{figure*}[t]
    \centering
    \includegraphics[width=\textwidth]{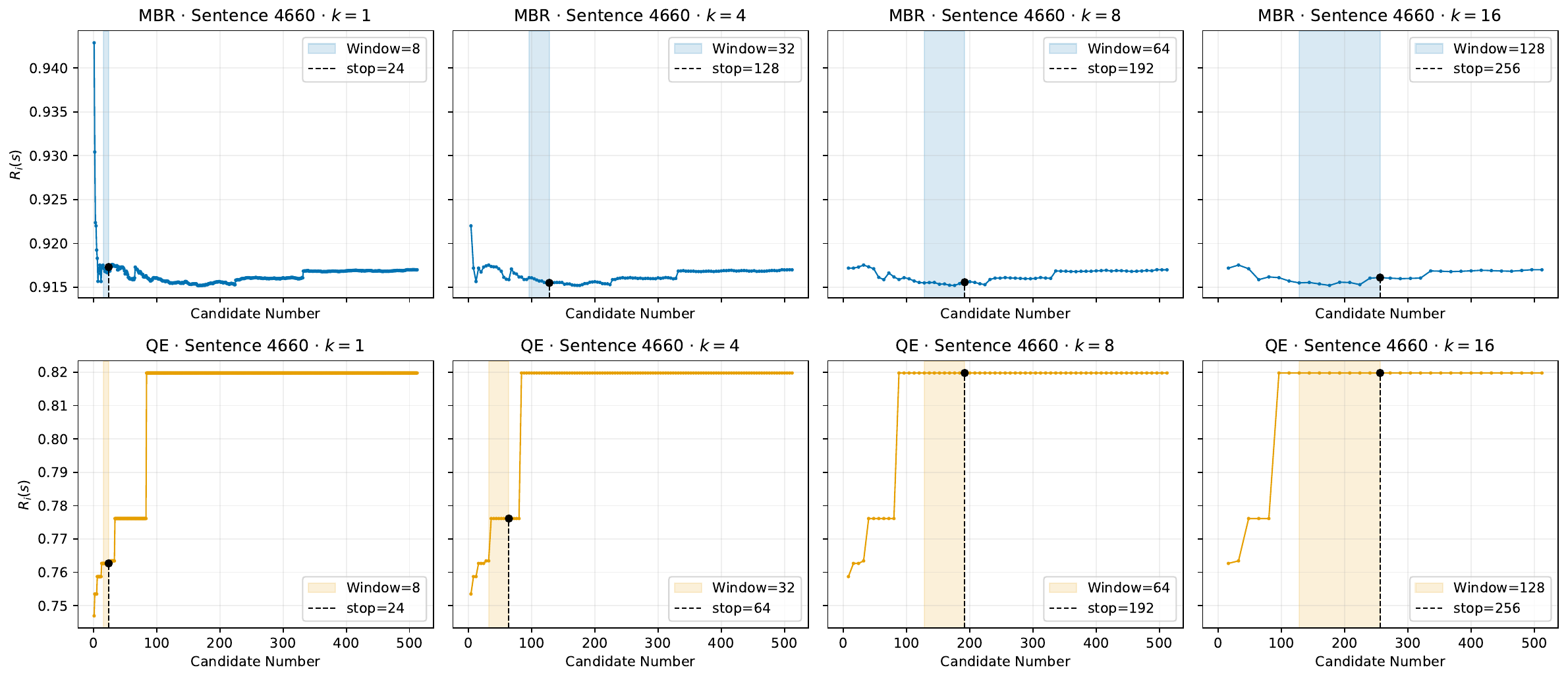}
    \caption{Example score trajectories for different batch sizes $k$ with a fixed window size $w=8$. A smaller $k$ may trigger early stopping too early, whereas a larger $k$ increases the candidate span of each window and can result in substantially more generated candidates before stopping.}
    \label{fig:batch_size_fixed_w}
\end{figure*}

With $w$ fixed, increasing $k$ enlarges the candidate span of each window and delays the earliest possible stopping point. 
As shown in Table~\ref{tab:batch_size_stopN}, the average stopping position consistently increases with $k$ for both MBR and QE. 
Figure~\ref{fig:batch_size_fixed_w} provides an example where a very small $k$ leads to early stopping, while a large $k$ results in substantially more generated candidates.

\begin{figure*}[t]
    \centering
    \includegraphics[width=\textwidth]{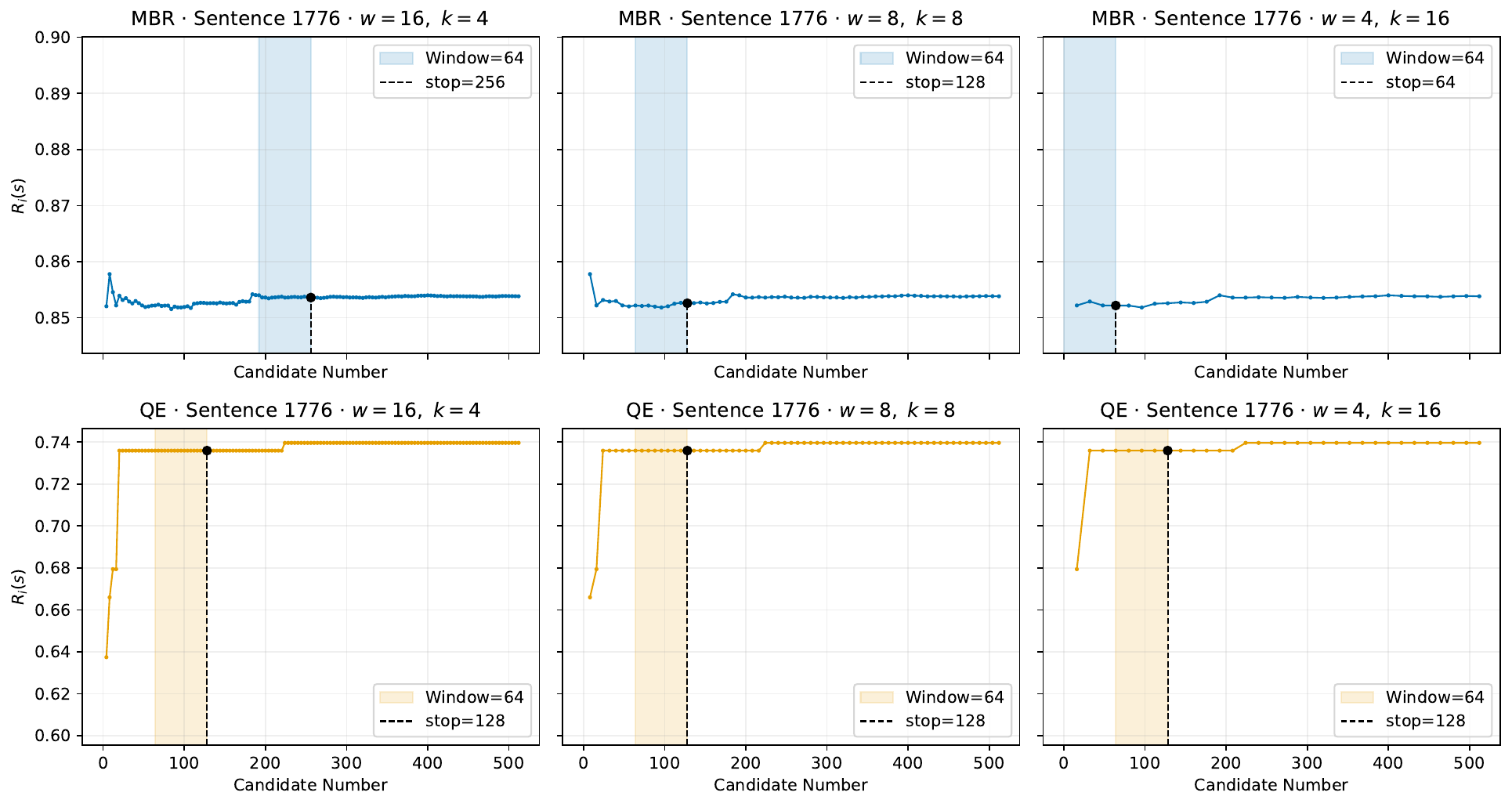}
    \caption{Example score trajectories for different $(k,w)$ combinations with a fixed candidate span $kw$. A smaller $k$ provides more frequent score observations within the same candidate span. This effect is more pronounced for MBR, whose scores can fluctuate as the support set expands, while QE is less affected because scores of existing candidates do not change with later candidates.}
    \label{fig:batch_size_fixed_kw}
\end{figure*}

When $kw$ is fixed, changing $k$ instead changes how frequently the score trajectory is observed within the same candidate span. 
This effect is more pronounced for MBR decoding. 
Since expanding the support set can change the scores of existing candidates, the MBR score may fluctuate locally; using a smaller $k$ gives more observations within the same span and is therefore more likely to capture such fluctuations, delaying stopping. 
In contrast, QE scores of existing candidates are unaffected by later candidates, so this effect is weaker. 
Figure~\ref{fig:batch_size_fixed_kw} illustrates this difference, while Table~\ref{tab:batch_size_stopN} shows the same trend on average over all evaluation data. 
This behavior is consistent with the different score dynamics of MBR and QE discussed in Appendix~\ref{sec:score_evolution}.

\section{Convergence-threshold Ablation}
\label{sec:alpha_ablation}

We evaluate $\alpha\in\{10^{-6},10^{-5},10^{-4},10^{-3},10^{-2}\}$ with $w\in\{4,8,16\}$ on the complete WMT24 and WMT25 test sets, while keeping all other settings unchanged from Section~\ref{sec:exp_setup}.
We report the mean number of generated candidates, the proportion of source sentences for which generation reaches the full candidate budget of 512, end-to-end speedup, and the same reranking and translation quality metrics as in the main experiments.

Tables~\ref{tab:alpha_mbr_w4}--\ref{tab:alpha_qe_w16} show macro-averages over the six test-set--model combinations.  We use paired-bootstrap tests with $10{,}000$ resamples against both the default $\alpha=10^{-3}$ and the unaccelerated baseline.
The final column counts how often the tested configuration performs significantly worse or better than the default configuration across the 30 test-set--model--metric comparisons.
MetricX is sign-reversed for testing but reported in its original lower-is-better orientation.
``Rerank.'' denotes the COMET-22 score for MBR and the CometKiwi-22 score for QE.

For MBR, $\alpha=10^{-3}$ marks a turning point in the trade-off between speedup and translation quality.
Tightening it by one order of magnitude causes generation to reach the full candidate budget for 42\%, 83\%, and 97\% of source sentences for $w=4$, $8$, and $16$, respectively, and reduces speedup to $1.43\times$, $1.06\times$, and $1.01\times$, with a significant improvement over the unaccelerated baseline in at most 3 of the 30 comparisons.

QE is less sensitive to small positive convergence thresholds.
For QE, the best reranking score $R_i(s)$ is a running maximum.
If it does not increase within a window, the uncertainty score $\Delta_b^{R}(s)$ is exactly zero, so the stopping criterion is satisfied for any $\alpha>0$.
Consequently, even $\alpha=10^{-6}$ retains much of the speedup.
A detectable but small degradation, approximately $-0.006\sigma$ on average, emerges only when the threshold is relaxed to $10^{-2}$.
Overall, QE is relatively insensitive to convergence thresholds below the default, whereas MBR shows a stronger trade-off between speedup and translation quality.

\section{Complete Window-size Sweep}
\label{sec:complete_window_sweep}

\begin{figure*}[t]
    \centering
    \includegraphics[width=\textwidth]{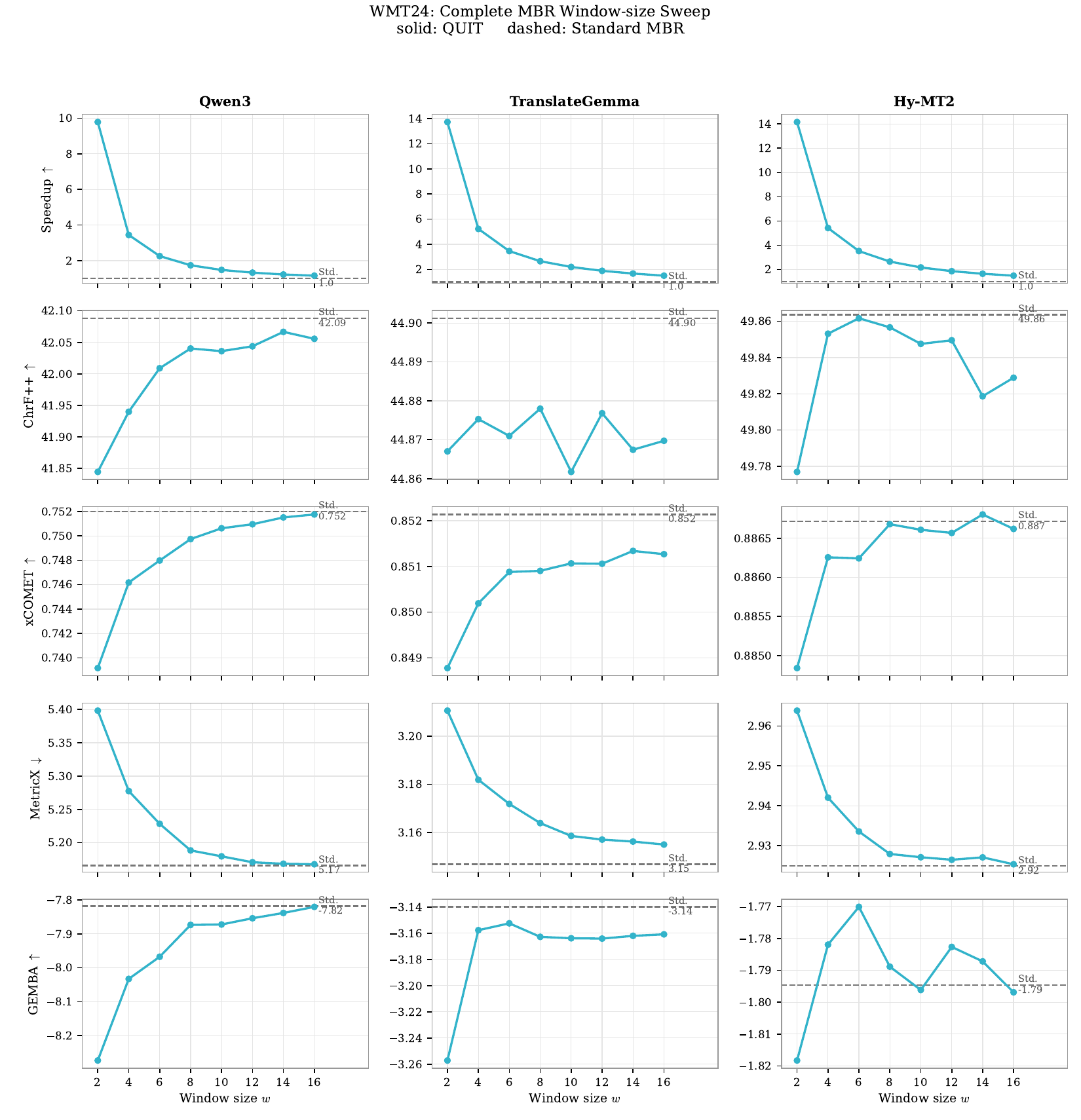}
    \caption{Complete MBR window-size sweep on WMT24.  Gray dashed lines mark the
    per-model unaccelerated MBR performance; the gray dashed line in the speedup
    panel marks the $1\times$ unaccelerated baseline.}
    \label{fig:window_sweep_wmt24_mbr}
\end{figure*}

\begin{figure*}[t]
    \centering
    \includegraphics[width=\textwidth]{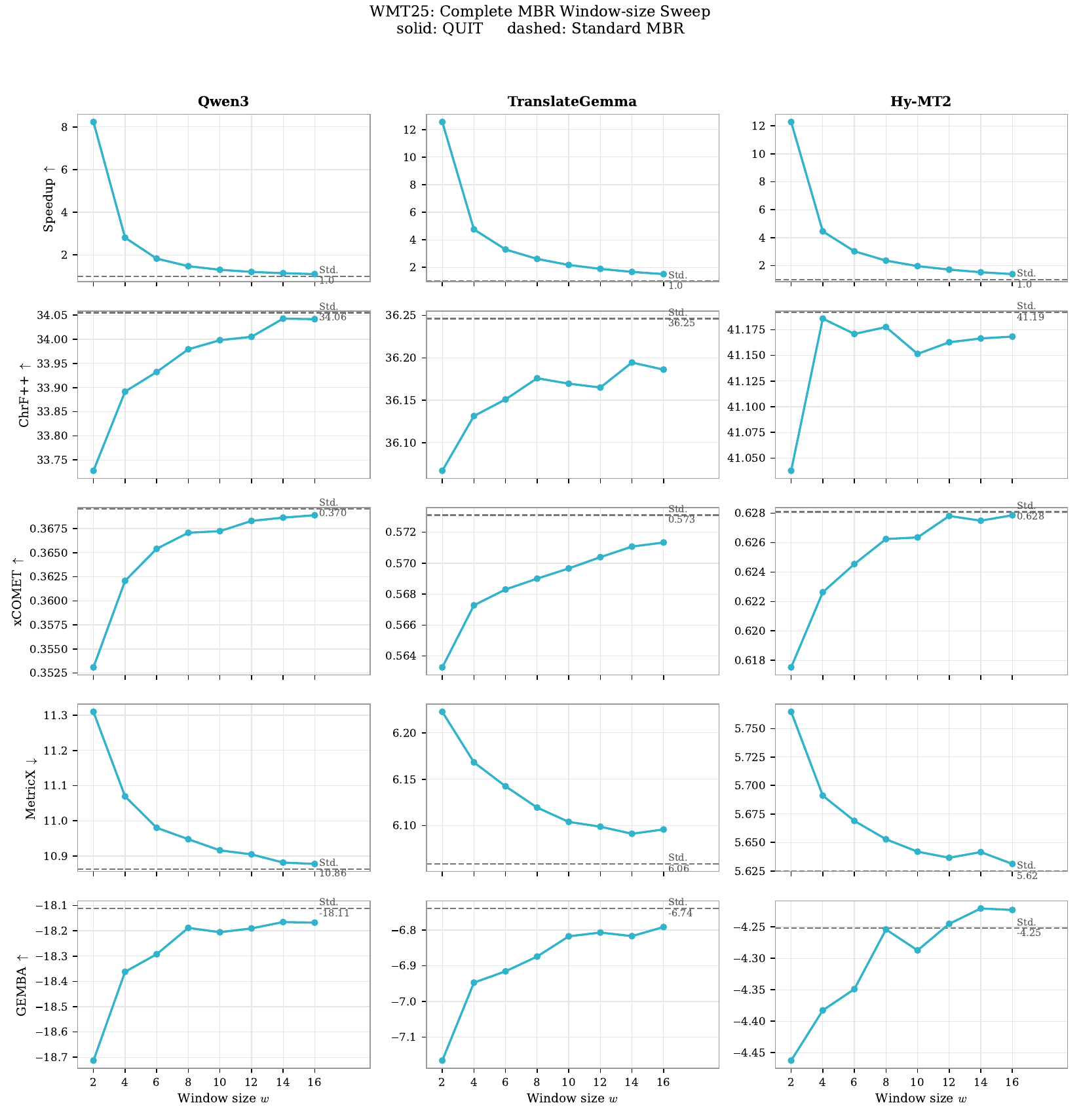}
    \caption{Complete MBR window-size sweep on WMT25, following the conventions
    of Figure~\ref{fig:window_sweep_wmt24_mbr}.}
    \label{fig:window_sweep_wmt25_mbr}
\end{figure*}

\begin{figure*}[t]
    \centering
    \includegraphics[width=\textwidth]{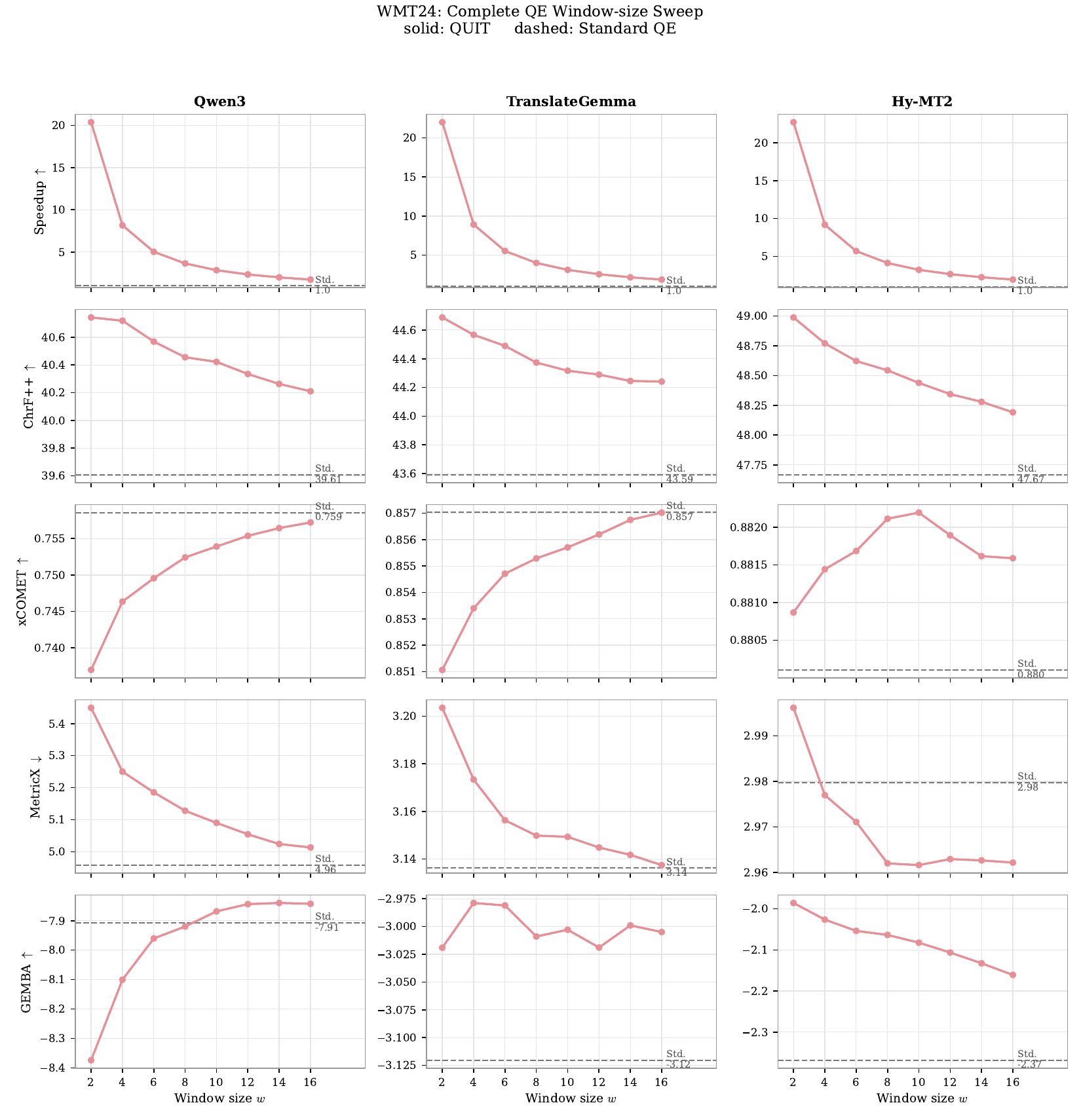}
    \caption{Complete QE window-size sweep on WMT24.  Gray dashed lines mark the
    per-model unaccelerated QE performance; the gray dashed line in the speedup
    panel marks the $1\times$ unaccelerated baseline.}
    \label{fig:window_sweep_wmt24_qe}
\end{figure*}

\begin{figure*}[t]
    \centering
    \includegraphics[width=\textwidth]{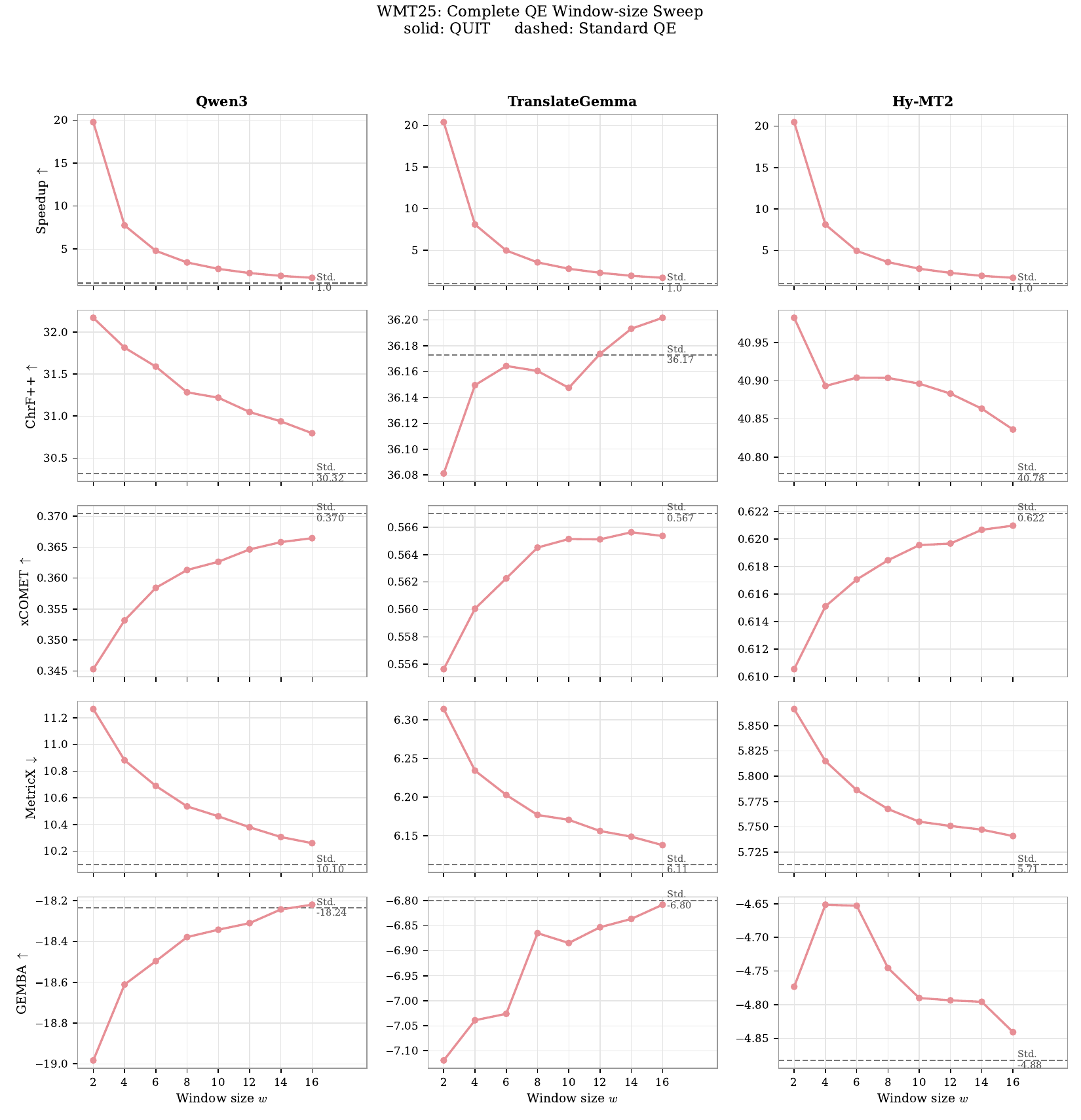}
    \caption{Complete QE window-size sweep on WMT25, following the conventions
    of Figure~\ref{fig:window_sweep_wmt24_qe}.}
    \label{fig:window_sweep_wmt25_qe}
\end{figure*}

Figures~\ref{fig:window_sweep_wmt24_mbr}--\ref{fig:window_sweep_wmt25_qe} report the complete sweep over $w\in\{2,4,6,8,10,12,14,16\}$ for every test set, NMT model, and reranking paradigm.  We show end-to-end speedup and all reference-based evaluation metrics reported in the main tables, while omitting the reranking score.
Solid lines show \textsc{Quit}; gray dashed lines show the corresponding unaccelerated MBR or QE baseline.  The speedup baseline is $1\times$ in every panel.  
Tables~\ref{tab:sweep_wmt24} and~\ref{tab:sweep_wmt25} list the corresponding numbers, with equivalence marks as in the main tables, for the window sizes not shown there.

\section{ChrF as the MBR Utility}
\label{sec:chrf_utility}

To verify that \textsc{Quit} remains effective with other utility functions, we repeat the MBR experiments with ChrF \cite{chrf} as the utility. Since ChrF is defined on a $0$--$100$ scale, we rescale it to $[0,1]$ before applying the stopping criterion. All other settings follow Section~\ref{sec:exp}: the maximum budget is $N_{\max}=512$, the convergence threshold is $\alpha=10^{-3}$, and we sweep $w \in \{2,4,\dots,16\}$. 
We evaluate with the same four external metrics and apply the same paired-bootstrap TOST procedure against the unaccelerated baseline, with equivalence margins of $0.05\sigma$ and $0.02\sigma$.

Table~\ref{tab:window_quality_results} summarizes the results.
Compared with the neural COMET-22 utility, ChrF is more sensitive to surface-form variation, leading to larger fluctuations in the $\{R_i(s)\}$ trajectories and slower convergence. 
Nevertheless, \textsc{Quit} remains effective: with $w=4$, all three models on WMT24 pass the stricter $0.02\sigma$ equivalence test on all four external metrics while achieving $2.48\times$--$3.08\times$ speedups. 
These results demonstrate that \textsc{Quit} is not tied to a specific utility function and remains effective across utilities with substantially different scoring characteristics.

\section{PRR Evaluation}
\label{sec:prr_computation}
We evaluate the full test sets, replaying the 512-candidate trajectories with $k=w=8$.
Each source--window pair $(s,b)$ is one instance, for $b=1,\ldots,7$, corresponding to candidate counts 64 through 448.
We exclude the final window because it already uses the full candidate pool and thus has zero early-stopping risk by construction.
For each source sentence, we include all seven windows, even those after the point where \textsc{Quit} would stop.
We then compute one PRR value for each test set, NMT model, reranker, and quality metric, using the source--window pairs from all evaluated sentences.

The risk is $\mathcal{L}_{bw}(s)$: the absolute difference between the quality of the output selected at step $bw$ and that of the output selected from all 512 candidates by the same reranker.
We compute it separately with xCOMET and GEMBA.
The uncertainty score is $\Delta_b^{R}(s)$, the within-window range of the reranking scores.

We progressively reject instances in descending uncertainty order and record the mean risk among the retained instances.
The oracle curve instead rejects instances in descending risk order, while the expected random-ranking curve stays at the overall mean risk.
We integrate these curves and compute PRR with Equation~\ref{eq:prr}.

Table~\ref{tab:prr_proxy_w8} evaluates $\Delta_b^{R}(s)$ against the early-stopping risk proxy $\mathcal{L}_{bw}^{\mathrm{proxy}}(s)$ from Equation~\ref{eq:proxy_early_stopping_risk}, using the same window setting as Table~\ref{tab:prr_w8}.
Here, the risk is computed from reranking scores, and instances are pooled within each test-set--model--reranker combination.
All values are positive, with mean PRR of 0.726 for MBR and 0.259 for QE.
These results support local reranking-score variation as an informative, imperfect indicator of the early-stopping risk proxy.

Tables~\ref{tab:prr_w4} and~\ref{tab:prr_w16} repeat the $\Delta_b^{R}(s)$ evaluation against $\mathcal{L}_{bw}(s)$ for the other two window sizes of the main tables ($w=4$ and $w=16$, i.e., $15$ and $3$ windows per source).

\section{Evolution of the Best Reranking Score}
\label{sec:score_evolution}

MBR and QE differ in their scoring mechanisms. 
MBR uses the average utility over a support set as a Monte Carlo approximation to expected utility as defined in Equation~\ref{eq:mbr}. 
As the candidate and support sets expand together, existing candidates' scores also change. 
QE, by contrast, scores each candidate independently, so its score is unaffected by later candidates according to Equation~\ref{eq:qe}. 
\textsc{Quit} tracks the best reranking score and stops generation when its within-window variation does not exceed a threshold, as defined in Equations~\ref{eq:best_reranking_score}–\ref{eq:stopping_time}.

\begin{figure*}[t]
      \centering
      \includegraphics[width=.9\textwidth]{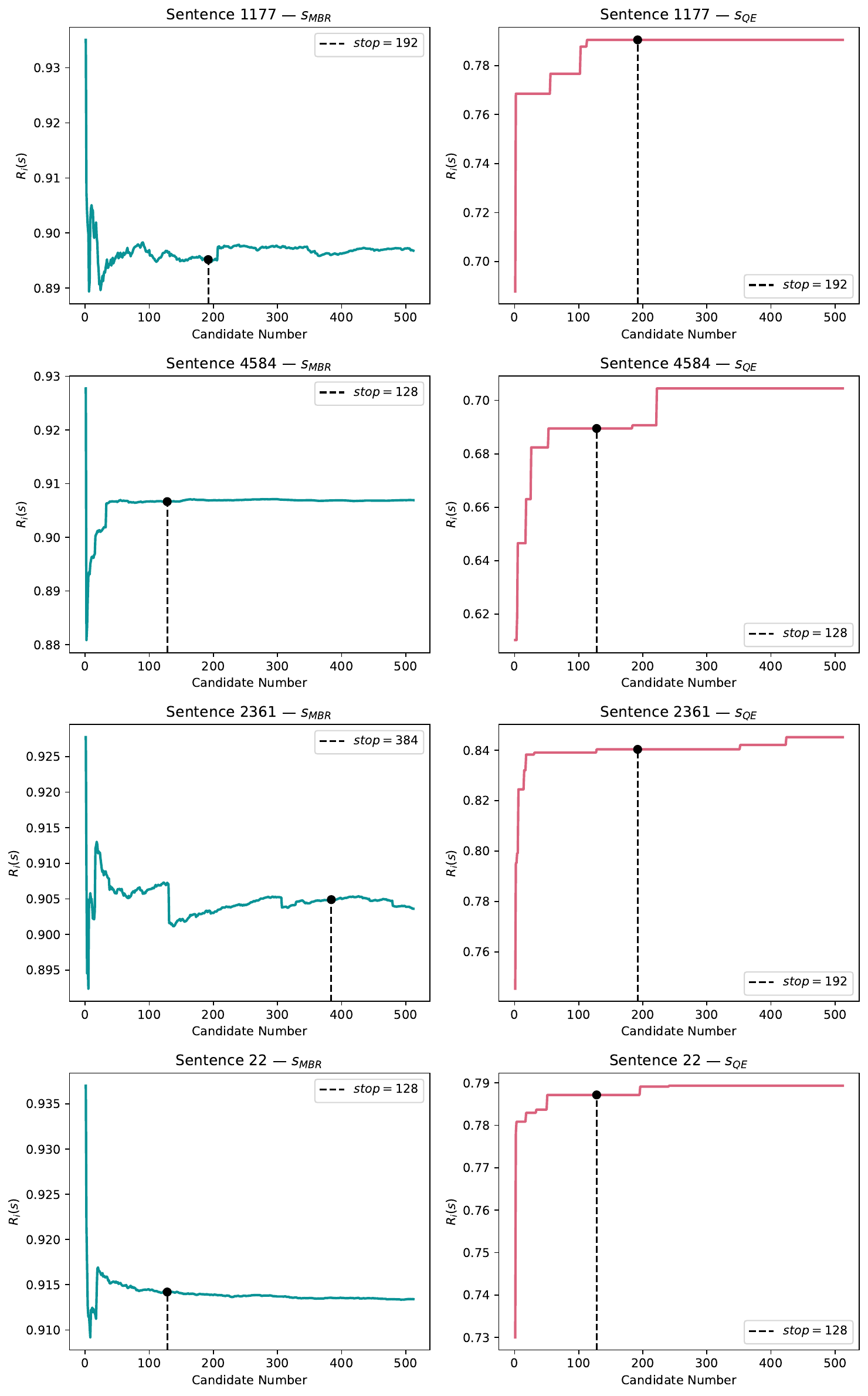}
      \caption{Evolution of the best reranking score $R_i(s)$ for four randomly selected source sentences from WMT25 with HY-MT2 as the candidate set grows. Each row corresponds to one source sentence, with MBR on the left and QE on the right. Black dashed lines and dots mark the stopping positions of \textsc{Quit} with $w=8$, $\alpha=0.001$, and $k=8$.}
      \label{fig:score_evolution}
\end{figure*}

Figure~\ref{fig:score_evolution} shows the score trajectories for four source sentences randomly sampled from WMT25-HY. 
Each row corresponds to one source sentence, with MBR on the left and QE on the right. 
MBR scores may fluctuate substantially early on because utility estimates based on small support sets are sensitive to additional samples, potentially delaying stabilization of the best score. 
The best QE score increases only when a higher-scoring candidate appears, producing a stepwise trajectory that more readily satisfies the stopping criterion.
These trajectories help explain why QE generally stops earlier, while MBR stopping positions are more broadly distributed.

\section{Component-wise Runtime}
\label{sec:runtime_details}

Tables~\ref{tab:runtime_wmt24} and~\ref{tab:runtime_wmt25} report the complete wall-clock measurements underlying Table~\ref{tab:runtime_breakdown}.
The Generation and Reranking columns report candidate-generation time and MBR reranking time, respectively, in seconds.
The aggregate measurements cover 12,010 source segments per model for WMT24 and 5,232 per model for WMT25; the per-segment averages in Table~\ref{tab:runtime_breakdown} pool all 51,726 segment--model runs.

% \section{Comparison with a Fixed-Budget Baseline}
% \label{sec:fixed_budget}

% We compare \textsc{Quit} at $w\in\{4,8,16\}$ and the default $\alpha=10^{-3}$ with a fixed-budget baseline, Fixed-$N^*$, which generates and reranks only the first $N^*$ candidates.
% For each test set, NMT model, reranker, and window size, we select $N^*$ from among multiples of eight to match \textsc{Quit}'s end-to-end cost as closely as possible.
% We otherwise follow the evaluation setup in Section~\ref{sec:exp_setup} and use paired-bootstrap 95\% confidence intervals to compare the two methods.

% Tables~\ref{tab:fixed_budget_w4}--\ref{tab:fixed_budget_w16} report the complete results.
% The two methods achieve nearly identical speedups by construction.
% Across most evaluation metrics, their quality differences are not significant; the significant differences that remain are small and show no consistent winner.

% This comparison favors Fixed-$N^*$ because $N^*$ is selected post hoc for each test set, NMT model, and reranker to match the cost of a \textsc{Quit} configuration that has passed the equivalence tests on nearly all external quality metrics.
% For each evaluated window size, \textsc{Quit} instead uses a shared $(w,\alpha)$ configuration and makes reference-free stopping decisions for each source sentence.
% Indeed, the oracle $N^*$ varies from 56 to 184 for $w=4$ and from 264 to 464 for $w=16$.
% These results suggest that a shared, adaptive stopping criterion can replace reference-dependent, setting-specific candidate-budget tuning.

\clearpage
\begin{table*}[t]
\centering
\small
\setlength{\tabcolsep}{5pt}
\begin{tabular}{@{}lcccccc@{}}
\toprule
\multirow{2}{*}{$\boldsymbol{k}$} &
\multicolumn{3}{c}{\textbf{MBR}} &
\multicolumn{3}{c}{\textbf{QE}} \\
\cmidrule(lr){2-4}\cmidrule(lr){5-7}
& $\boldsymbol{w=4}$ & $\boldsymbol{w=8}$ & $\boldsymbol{w=16}$
& $\boldsymbol{w=4}$ & $\boldsymbol{w=8}$ & $\boldsymbol{w=16}$ \\
\midrule
4  & 81.11  & 162.55 & 271.38 & 29.05  & 65.04  & 138.53 \\
8  & 132.86 & 252.12 & 388.65 & 57.43  & 128.62 & 272.71 \\
16 & 214.76 & 371.20 & 501.06 & 113.16 & 253.46 & 461.15 \\
\bottomrule
\end{tabular}
\caption{Average number of generated candidates at stopping for different generation batch sizes $k$ and window sizes $w$, averaged over all data.}
\label{tab:batch_size_stopN}
\end{table*}

\clearpage
\begin{table*}[t]
\centering
\small
\setlength{\tabcolsep}{4pt}
%\resizebox{\textwidth}{!}{%
\begin{tabular}{lrrrrrrrrr}
\toprule
$\boldsymbol{\alpha}$ & $\boldsymbol{\overline{N}}$ & \textbf{Full (\%)} &
\textbf{Speedup} & \textbf{Rerank.} & \textbf{ChrF++} & \textbf{xCOMET} &
\textbf{MetricX} & \textbf{GEMBA} & $\boldsymbol{n_{\downarrow}/n_{\uparrow}}$ \\
\midrule
$10^{-6}$ & 501 & 97.6 & 1.00 & .8239 & 41.39 & .6770 & 5.63 & $-6.96$ & 0/24 \\
$10^{-5}$ & 499 & 96.4 & 1.01 & .8239 & 41.39 & .6770 & 5.63 & $-6.96$ & 0/24 \\
$10^{-4}$ & 380 & 42.5 & 1.43 & .8236 & 41.36 & .6761 & 5.65 & $-7.00$ & 0/21 \\
$\mathbf{10^{-3}}$ & \textbf{133} & \textbf{1.2} & \textbf{4.35} & \textbf{.8209} & \textbf{41.31} & \textbf{.6724} & \textbf{5.72} & $\mathbf{-7.11}$ & -- \\
$10^{-2}$ & 45 & 0.0 & 12.19 & .8173 & 41.24 & .6682 & 5.81 & $-7.28$ & 26/0 \\
\bottomrule
\end{tabular}%
%}
\caption{$\alpha$ ablation for MBR with $w=4$.  ``Full (\%)'' is the percentage of source sentences that reach 512 candidates; $n_{\downarrow}$ and $n_{\uparrow}$ count comparisons in which the tested configuration performs significantly worse and better than the default, respectively.}
\label{tab:alpha_mbr_w4}
\end{table*}

\begin{table*}[t]
\centering
\small
\setlength{\tabcolsep}{4pt}
%\resizebox{\textwidth}{!}{%
\begin{tabular}{lrrrrrrrrr}
\toprule
$\boldsymbol{\alpha}$ & $\boldsymbol{\overline{N}}$ & \textbf{Full (\%)} &
\textbf{Speedup} & \textbf{Rerank.} & \textbf{ChrF++} & \textbf{xCOMET} &
\textbf{MetricX} & \textbf{GEMBA} & $\boldsymbol{n_{\downarrow}/n_{\uparrow}}$ \\
\midrule
$10^{-6}$ & 504 & 98.1 & 1.00 & .8239 & 41.39 & .6770 & 5.63 & $-6.96$ & 0/20 \\
$10^{-5}$ & 504 & 98.1 & 1.00 & .8239 & 41.39 & .6770 & 5.63 & $-6.96$ & 0/20 \\
$10^{-4}$ & 481 & 82.6 & 1.06 & .8239 & 41.38 & .6770 & 5.63 & $-6.96$ & 0/20 \\
$\mathbf{10^{-3}}$ & \textbf{252} & \textbf{12.1} & \textbf{2.25} & \textbf{.8227} & \textbf{41.35} & \textbf{.6749} & \textbf{5.67} & $\mathbf{-7.02}$ & -- \\
$10^{-2}$ & 95 & 0.0 & 5.83 & .8198 & 41.30 & .6716 & 5.74 & $-7.14$ & 21/0 \\
\bottomrule
\end{tabular}%
%}
\caption{$\alpha$ ablation for MBR with $w=8$, following the conventions of Table~\ref{tab:alpha_mbr_w4}.}
\label{tab:alpha_mbr_w8}
\end{table*}

\begin{table*}[t]
\centering
\small
\setlength{\tabcolsep}{4pt}
%\resizebox{\textwidth}{!}{%
\begin{tabular}{lrrrrrrrrr}
\toprule
$\boldsymbol{\alpha}$ & $\boldsymbol{\overline{N}}$ & \textbf{Full (\%)} &
\textbf{Speedup} & \textbf{Rerank.} & \textbf{ChrF++} & \textbf{xCOMET} &
\textbf{MetricX} & \textbf{GEMBA} & $\boldsymbol{n_{\downarrow}/n_{\uparrow}}$ \\
\midrule
$10^{-6}$ & 506 & 98.5 & 1.00 & .8239 & 41.39 & .6770 & 5.63 & $-6.96$ & 0/16 \\
$10^{-5}$ & 506 & 98.5 & 1.00 & .8239 & 41.39 & .6770 & 5.63 & $-6.96$ & 0/16 \\
$10^{-4}$ & 505 & 97.4 & 1.01 & .8239 & 41.39 & .6770 & 5.63 & $-6.96$ & 0/16 \\
$\mathbf{10^{-3}}$ & \textbf{389} & \textbf{40.8} & \textbf{1.36} & \textbf{.8236} & \textbf{41.36} & \textbf{.6763} & \textbf{5.64} & $\mathbf{-6.99}$ & -- \\
$10^{-2}$ & 191 & 0.6 & 2.86 & .8220 & 41.35 & .6744 & 5.69 & $-7.04$ & 18/0 \\
\bottomrule
\end{tabular}%
%}
\caption{$\alpha$ ablation for MBR with $w=16$, following the conventions of Table~\ref{tab:alpha_mbr_w4}.}
\label{tab:alpha_mbr_w16}
\end{table*}

\begin{table*}[t]
\centering
\small
\setlength{\tabcolsep}{4pt}
%\resizebox{\textwidth}{!}{%
\begin{tabular}{lrrrrrrrrr}
\toprule
$\boldsymbol{\alpha}$ & $\boldsymbol{\overline{N}}$ & \textbf{Full (\%)} &
\textbf{Speedup} & \textbf{Rerank.} & \textbf{ChrF++} & \textbf{xCOMET} &
\textbf{MetricX} & \textbf{GEMBA} & $\boldsymbol{n_{\downarrow}/n_{\uparrow}}$ \\
\midrule
$10^{-6}$ & 64 & 0.0 & 7.73 & .7795 & 40.48 & .6685 & 5.72 & $-7.23$ & 0/8 \\
$10^{-5}$ & 64 & 0.0 & 7.74 & .7795 & 40.48 & .6685 & 5.72 & $-7.23$ & 0/9 \\
$10^{-4}$ & 63 & 0.0 & 7.81 & .7795 & 40.48 & .6685 & 5.72 & $-7.24$ & 0/11 \\
$\mathbf{10^{-3}}$ & \textbf{59} & \textbf{0.0} & \textbf{8.37} & \textbf{.7793} & \textbf{40.49} & \textbf{.6683} & \textbf{5.72} & $\mathbf{-7.23}$ & -- \\
$10^{-2}$ & 45 & 0.0 & 11.46 & .7777 & 40.49 & .6669 & 5.75 & $-7.28$ & 17/1 \\
\bottomrule
\end{tabular}%
%}
\caption{$\alpha$ ablation for QE with $w=4$, following the conventions of Table~\ref{tab:alpha_mbr_w4}.}
\label{tab:alpha_qe_w4}
\end{table*}

\begin{table*}[t]
\centering
\small
\setlength{\tabcolsep}{4pt}
%\resizebox{\textwidth}{!}{%
\begin{tabular}{lrrrrrrrrr}
\toprule
$\boldsymbol{\alpha}$ & $\boldsymbol{\overline{N}}$ & \textbf{Full (\%)} &
\textbf{Speedup} & \textbf{Rerank.} & \textbf{ChrF++} & \textbf{xCOMET} &
\textbf{MetricX} & \textbf{GEMBA} & $\boldsymbol{n_{\downarrow}/n_{\uparrow}}$ \\
\midrule
$10^{-6}$ & 144 & 0.0 & 3.41 & .7852 & 40.29 & .6727 & 5.62 & $-7.16$ & 0/8 \\
$10^{-5}$ & 144 & 0.0 & 3.41 & .7852 & 40.29 & .6727 & 5.62 & $-7.16$ & 0/8 \\
$10^{-4}$ & 142 & 0.0 & 3.45 & .7852 & 40.29 & .6726 & 5.62 & $-7.16$ & 1/8 \\
$\mathbf{10^{-3}}$ & \textbf{133} & \textbf{0.0} & \textbf{3.72} & \textbf{.7849} & \textbf{40.29} & \textbf{.6723} & \textbf{5.62} & $\mathbf{-7.16}$ & -- \\
$10^{-2}$ & 97 & 0.0 & 5.25 & .7834 & 40.32 & .6711 & 5.65 & $-7.18$ & 15/2 \\
\bottomrule
\end{tabular}%
%}
\caption{$\alpha$ ablation for QE with $w=8$, following the conventions of Table~\ref{tab:alpha_mbr_w4}.}
\label{tab:alpha_qe_w8}
\end{table*}

\begin{table*}[t]
\centering
\small
\setlength{\tabcolsep}{4pt}
%\resizebox{\textwidth}{!}{%
\begin{tabular}{lrrrrrrrrr}
\toprule
$\boldsymbol{\alpha}$ & $\boldsymbol{\overline{N}}$ & \textbf{Full (\%)} &
\textbf{Speedup} & \textbf{Rerank.} & \textbf{ChrF++} & \textbf{xCOMET} &
\textbf{MetricX} & \textbf{GEMBA} & $\boldsymbol{n_{\downarrow}/n_{\uparrow}}$ \\
\midrule
$10^{-6}$ & 303 & 11.1 & 1.62 & .7898 & 40.08 & .6749 & 5.54 & $-7.15$ & 1/7 \\
$10^{-5}$ & 303 & 11.1 & 1.62 & .7898 & 40.08 & .6749 & 5.54 & $-7.15$ & 1/8 \\
$10^{-4}$ & 300 & 10.4 & 1.64 & .7898 & 40.08 & .6749 & 5.54 & $-7.15$ & 2/7 \\
$\mathbf{10^{-3}}$ & \textbf{281} & \textbf{6.6} & \textbf{1.76} & \textbf{.7896} & \textbf{40.08} & \textbf{.6748} & \textbf{5.54} & $\mathbf{-7.15}$ & -- \\
$10^{-2}$ & 206 & 0.7 & 2.48 & .7882 & 40.11 & .6740 & 5.56 & $-7.14$ & 12/2 \\
\bottomrule
\end{tabular}%
%}
\caption{$\alpha$ ablation for QE with $w=16$, following the conventions of Table~\ref{tab:alpha_mbr_w4}.}
\label{tab:alpha_qe_w16}
\end{table*}

\clearpage

\begin{table*}[t]
\centering
\footnotesize
\setlength{\tabcolsep}{3.5pt}
\begin{tabular}{@{}lllcccccc@{}}
\toprule
\textbf{NMT Model} & \textbf{Reranker} & \textbf{Acceleration} & \textbf{Speedup $\uparrow$} & \textbf{Rerank. $\uparrow$} & \textbf{ChrF++ $\uparrow$} & \textbf{xCOMET $\uparrow$} & \textbf{MetricX $\downarrow$} & \textbf{GEMBA $\uparrow$} \\
\midrule
\multirow{14}{*}{Qwen3} & \multirow{8}{*}{MBR} & Unaccelerated & 1.00 & .8357 & 42.09 & .7520 & 5.17 & -7.82 \\
 &  & PruneMBR & 1.05 & ~~~.8355\textsuperscript{\dag\dag} & ~~~42.07\textsuperscript{\dag\dag} & ~~~.7519\textsuperscript{\dag\dag} & ~~~5.17\textsuperscript{\dag\dag} & ~~~-7.81\textsuperscript{\dag\dag} \\
 &  & PMBR & 1.03 & ~~~.8352\textsuperscript{\dag\dag} & ~~~42.11\textsuperscript{\dag\dag} & ~~~.7516\textsuperscript{\dag\dag} & ~~~5.17\textsuperscript{\dag\dag} & ~~~-7.81\textsuperscript{\dag\dag} \\
 &  & \textsc{Quit} ($w=2$) & 9.78 & .8255 & ~~~41.84\textsuperscript{\dag\dag} & .7392 & 5.40 & -8.27 \\
 &  & \textsc{Quit} ($w=4$) & 3.44 & ~.8315\textsuperscript{\dag} & ~~~41.94\textsuperscript{\dag\dag} & ~.7462\textsuperscript{\dag} & ~5.28\textsuperscript{\dag} & ~-8.03\textsuperscript{\dag} \\
 &  & \textsc{Quit} ($w=6$) & 2.26 & ~.8334\textsuperscript{\dag} & ~~~42.01\textsuperscript{\dag\dag} & ~.7480\textsuperscript{\dag} & ~~~5.23\textsuperscript{\dag\dag} & ~-7.97\textsuperscript{\dag} \\
 &  & \textsc{Quit} ($w=8$) & 1.74 & ~~~.8344\textsuperscript{\dag\dag} & ~~~42.04\textsuperscript{\dag\dag} & ~~~.7497\textsuperscript{\dag\dag} & ~~~5.19\textsuperscript{\dag\dag} & ~~~-7.87\textsuperscript{\dag\dag} \\
 &  & \textsc{Quit} ($w=10$) & 1.48 & ~~~.8349\textsuperscript{\dag\dag} & ~~~42.04\textsuperscript{\dag\dag} & ~~~.7506\textsuperscript{\dag\dag} & ~~~5.18\textsuperscript{\dag\dag} & ~~~-7.87\textsuperscript{\dag\dag} \\
 &  & \textsc{Quit} ($w=12$) & 1.33 & ~~~.8352\textsuperscript{\dag\dag} & ~~~42.04\textsuperscript{\dag\dag} & ~~~.7510\textsuperscript{\dag\dag} & ~~~5.17\textsuperscript{\dag\dag} & ~~~-7.85\textsuperscript{\dag\dag} \\
 &  & \textsc{Quit} ($w=14$) & 1.22 & ~~~.8354\textsuperscript{\dag\dag} & ~~~42.07\textsuperscript{\dag\dag} & ~~~.7515\textsuperscript{\dag\dag} & ~~~5.17\textsuperscript{\dag\dag} & ~~~-7.84\textsuperscript{\dag\dag} \\
 &  & \textsc{Quit} ($w=16$) & 1.16 & ~~~.8355\textsuperscript{\dag\dag} & ~~~42.06\textsuperscript{\dag\dag} & ~~~.7518\textsuperscript{\dag\dag} & ~~~5.17\textsuperscript{\dag\dag} & ~~~-7.82\textsuperscript{\dag\dag} \\
\cmidrule{2-9}
 & \multirow{6}{*}{QE} & Unaccelerated & 1.00 & .8223 & 39.61 & .7585 & 4.96 & -7.91 \\
 &  & \textsc{Quit} ($w=2$) & 20.37 & .7975 & 40.74 & .7370 & 5.45 & -8.37 \\
 &  & \textsc{Quit} ($w=4$) & 8.16 & .8067 & 40.72 & .7464 & 5.25 & ~-8.10\textsuperscript{\dag} \\
 &  & \textsc{Quit} ($w=6$) & 5.02 & .8108 & 40.57 & ~.7496\textsuperscript{\dag} & 5.19 & ~~~-7.96\textsuperscript{\dag\dag} \\
 &  & \textsc{Quit} ($w=8$) & 3.64 & .8133 & ~40.46\textsuperscript{\dag} & ~.7524\textsuperscript{\dag} & ~5.13\textsuperscript{\dag} & ~~~-7.92\textsuperscript{\dag\dag} \\
 &  & \textsc{Quit} ($w=10$) & 2.84 & .8151 & ~40.42\textsuperscript{\dag} & ~.7539\textsuperscript{\dag} & ~5.09\textsuperscript{\dag} & ~~~-7.87\textsuperscript{\dag\dag} \\
 &  & \textsc{Quit} ($w=12$) & 2.33 & .8165 & ~40.33\textsuperscript{\dag} & ~~~.7554\textsuperscript{\dag\dag} & ~5.05\textsuperscript{\dag} & ~~~-7.85\textsuperscript{\dag\dag} \\
 &  & \textsc{Quit} ($w=14$) & 1.99 & .8175 & ~40.26\textsuperscript{\dag} & ~~~.7564\textsuperscript{\dag\dag} & ~~~5.02\textsuperscript{\dag\dag} & ~~~-7.84\textsuperscript{\dag\dag} \\
 &  & \textsc{Quit} ($w=16$) & 1.73 & ~.8185\textsuperscript{\dag} & ~40.21\textsuperscript{\dag} & ~~~.7572\textsuperscript{\dag\dag} & ~~~5.01\textsuperscript{\dag\dag} & ~~~-7.84\textsuperscript{\dag\dag} \\
\midrule
\multirow{14}{*}{TranslateGemma} & \multirow{8}{*}{MBR} & Unaccelerated & 1.00 & .8570 & 44.90 & .8521 & 3.15 & -3.14 \\
 &  & PruneMBR & 1.04 & ~~~.8570\textsuperscript{\dag\dag} & ~~~44.88\textsuperscript{\dag\dag} & ~~~.8520\textsuperscript{\dag\dag} & ~~~3.15\textsuperscript{\dag\dag} & ~~~-3.13\textsuperscript{\dag\dag} \\
 &  & PMBR & 1.03 & ~~~.8566\textsuperscript{\dag\dag} & ~~~44.91\textsuperscript{\dag\dag} & ~~~.8518\textsuperscript{\dag\dag} & ~~~3.15\textsuperscript{\dag\dag} & ~~~-3.16\textsuperscript{\dag\dag} \\
 &  & \textsc{Quit} ($w=2$) & 13.72 & ~.8543\textsuperscript{\dag} & ~~~44.87\textsuperscript{\dag\dag} & ~.8488\textsuperscript{\dag} & ~3.21\textsuperscript{\dag} & ~-3.26\textsuperscript{\dag} \\
 &  & \textsc{Quit} ($w=4$) & 5.23 & ~~~.8557\textsuperscript{\dag\dag} & ~~~44.88\textsuperscript{\dag\dag} & ~~~.8502\textsuperscript{\dag\dag} & ~~~3.18\textsuperscript{\dag\dag} & ~~~-3.16\textsuperscript{\dag\dag} \\
 &  & \textsc{Quit} ($w=6$) & 3.46 & ~~~.8563\textsuperscript{\dag\dag} & ~~~44.87\textsuperscript{\dag\dag} & ~~~.8509\textsuperscript{\dag\dag} & ~~~3.17\textsuperscript{\dag\dag} & ~~~-3.15\textsuperscript{\dag\dag} \\
 &  & \textsc{Quit} ($w=8$) & 2.66 & ~~~.8566\textsuperscript{\dag\dag} & ~~~44.88\textsuperscript{\dag\dag} & ~~~.8509\textsuperscript{\dag\dag} & ~~~3.16\textsuperscript{\dag\dag} & ~~~-3.16\textsuperscript{\dag\dag} \\
 &  & \textsc{Quit} ($w=10$) & 2.20 & ~~~.8567\textsuperscript{\dag\dag} & ~~~44.86\textsuperscript{\dag\dag} & ~~~.8511\textsuperscript{\dag\dag} & ~~~3.16\textsuperscript{\dag\dag} & ~~~-3.16\textsuperscript{\dag\dag} \\
 &  & \textsc{Quit} ($w=12$) & 1.89 & ~~~.8568\textsuperscript{\dag\dag} & ~~~44.88\textsuperscript{\dag\dag} & ~~~.8511\textsuperscript{\dag\dag} & ~~~3.16\textsuperscript{\dag\dag} & ~~~-3.16\textsuperscript{\dag\dag} \\
 &  & \textsc{Quit} ($w=14$) & 1.68 & ~~~.8569\textsuperscript{\dag\dag} & ~~~44.87\textsuperscript{\dag\dag} & ~~~.8513\textsuperscript{\dag\dag} & ~~~3.16\textsuperscript{\dag\dag} & ~~~-3.16\textsuperscript{\dag\dag} \\
 &  & \textsc{Quit} ($w=16$) & 1.51 & ~~~.8569\textsuperscript{\dag\dag} & ~~~44.87\textsuperscript{\dag\dag} & ~~~.8513\textsuperscript{\dag\dag} & ~~~3.15\textsuperscript{\dag\dag} & ~~~-3.16\textsuperscript{\dag\dag} \\
\cmidrule{2-9}
 & \multirow{6}{*}{QE} & Unaccelerated & 1.00 & .8376 & 43.59 & .8570 & 3.14 & -3.12 \\
 &  & \textsc{Quit} ($w=2$) & 22.01 & .8251 & 44.69 & ~.8511\textsuperscript{\dag} & ~3.20\textsuperscript{\dag} & ~-3.01\textsuperscript{\dag} \\
 &  & \textsc{Quit} ($w=4$) & 8.92 & .8295 & 44.57 & ~.8534\textsuperscript{\dag} & ~3.17\textsuperscript{\dag} & ~-2.98\textsuperscript{\dag} \\
 &  & \textsc{Quit} ($w=6$) & 5.54 & .8314 & 44.49 & ~.8547\textsuperscript{\dag} & ~~~3.16\textsuperscript{\dag\dag} & ~-2.98\textsuperscript{\dag} \\
 &  & \textsc{Quit} ($w=8$) & 4.01 & .8327 & ~44.37\textsuperscript{\dag} & ~~~.8553\textsuperscript{\dag\dag} & ~~~3.15\textsuperscript{\dag\dag} & ~-3.01\textsuperscript{\dag} \\
 &  & \textsc{Quit} ($w=10$) & 3.12 & .8336 & ~44.32\textsuperscript{\dag} & ~~~.8557\textsuperscript{\dag\dag} & ~~~3.15\textsuperscript{\dag\dag} & ~-3.00\textsuperscript{\dag} \\
 &  & \textsc{Quit} ($w=12$) & 2.57 & .8344 & ~44.29\textsuperscript{\dag} & ~~~.8562\textsuperscript{\dag\dag} & ~~~3.14\textsuperscript{\dag\dag} & ~-3.02\textsuperscript{\dag} \\
 &  & \textsc{Quit} ($w=14$) & 2.17 & .8349 & ~44.25\textsuperscript{\dag} & ~~~.8567\textsuperscript{\dag\dag} & ~~~3.14\textsuperscript{\dag\dag} & ~-3.00\textsuperscript{\dag} \\
 &  & \textsc{Quit} ($w=16$) & 1.88 & ~.8355\textsuperscript{\dag} & ~44.24\textsuperscript{\dag} & ~~~.8570\textsuperscript{\dag\dag} & ~~~3.14\textsuperscript{\dag\dag} & ~-3.00\textsuperscript{\dag} \\
\midrule
\multirow{14}{*}{Hy-MT2} & \multirow{8}{*}{MBR} & Unaccelerated & 1.00 & .8731 & 49.86 & .8867 & 2.92 & -1.79 \\
 &  & PruneMBR & 1.02 & ~~~.8732\textsuperscript{\dag\dag} & ~~~49.87\textsuperscript{\dag\dag} & ~~~.8869\textsuperscript{\dag\dag} & ~~~2.92\textsuperscript{\dag\dag} & ~~~-1.80\textsuperscript{\dag\dag} \\
 &  & PMBR & 1.02 & ~~~.8728\textsuperscript{\dag\dag} & ~~~49.87\textsuperscript{\dag\dag} & ~~~.8866\textsuperscript{\dag\dag} & ~~~2.93\textsuperscript{\dag\dag} & ~~~-1.79\textsuperscript{\dag\dag} \\
 &  & \textsc{Quit} ($w=2$) & 14.16 & ~.8712\textsuperscript{\dag} & ~~~49.78\textsuperscript{\dag\dag} & ~~~.8848\textsuperscript{\dag\dag} & ~2.96\textsuperscript{\dag} & ~~~-1.82\textsuperscript{\dag\dag} \\
 &  & \textsc{Quit} ($w=4$) & 5.42 & ~~~.8723\textsuperscript{\dag\dag} & ~~~49.85\textsuperscript{\dag\dag} & ~~~.8863\textsuperscript{\dag\dag} & ~~~2.94\textsuperscript{\dag\dag} & ~~~-1.78\textsuperscript{\dag\dag} \\
 &  & \textsc{Quit} ($w=6$) & 3.52 & ~~~.8726\textsuperscript{\dag\dag} & ~~~49.86\textsuperscript{\dag\dag} & ~~~.8862\textsuperscript{\dag\dag} & ~~~2.93\textsuperscript{\dag\dag} & ~~~-1.77\textsuperscript{\dag\dag} \\
 &  & \textsc{Quit} ($w=8$) & 2.66 & ~~~.8728\textsuperscript{\dag\dag} & ~~~49.86\textsuperscript{\dag\dag} & ~~~.8867\textsuperscript{\dag\dag} & ~~~2.93\textsuperscript{\dag\dag} & ~~~-1.79\textsuperscript{\dag\dag} \\
 &  & \textsc{Quit} ($w=10$) & 2.18 & ~~~.8729\textsuperscript{\dag\dag} & ~~~49.85\textsuperscript{\dag\dag} & ~~~.8866\textsuperscript{\dag\dag} & ~~~2.93\textsuperscript{\dag\dag} & ~~~-1.80\textsuperscript{\dag\dag} \\
 &  & \textsc{Quit} ($w=12$) & 1.87 & ~~~.8730\textsuperscript{\dag\dag} & ~~~49.85\textsuperscript{\dag\dag} & ~~~.8866\textsuperscript{\dag\dag} & ~~~2.93\textsuperscript{\dag\dag} & ~~~-1.78\textsuperscript{\dag\dag} \\
 &  & \textsc{Quit} ($w=14$) & 1.65 & ~~~.8730\textsuperscript{\dag\dag} & ~~~49.82\textsuperscript{\dag\dag} & ~~~.8868\textsuperscript{\dag\dag} & ~~~2.93\textsuperscript{\dag\dag} & ~~~-1.79\textsuperscript{\dag\dag} \\
 &  & \textsc{Quit} ($w=16$) & 1.50 & ~~~.8731\textsuperscript{\dag\dag} & ~~~49.83\textsuperscript{\dag\dag} & ~~~.8866\textsuperscript{\dag\dag} & ~~~2.93\textsuperscript{\dag\dag} & ~~~-1.80\textsuperscript{\dag\dag} \\
\cmidrule{2-9}
 & \multirow{6}{*}{QE} & Unaccelerated & 1.00 & .8366 & 47.67 & .8801 & 2.98 & -2.37 \\
 &  & \textsc{Quit} ($w=2$) & 22.70 & .8254 & 48.99 & ~~~.8809\textsuperscript{\dag\dag} & ~~~3.00\textsuperscript{\dag\dag} & -1.98 \\
 &  & \textsc{Quit} ($w=4$) & 9.17 & .8290 & 48.77 & ~~~.8814\textsuperscript{\dag\dag} & ~~~2.98\textsuperscript{\dag\dag} & -2.03 \\
 &  & \textsc{Quit} ($w=6$) & 5.70 & .8306 & ~48.62\textsuperscript{\dag} & ~~~.8817\textsuperscript{\dag\dag} & ~~~2.97\textsuperscript{\dag\dag} & -2.05 \\
 &  & \textsc{Quit} ($w=8$) & 4.12 & .8317 & ~48.54\textsuperscript{\dag} & ~~~.8821\textsuperscript{\dag\dag} & ~~~2.96\textsuperscript{\dag\dag} & -2.06 \\
 &  & \textsc{Quit} ($w=10$) & 3.22 & .8325 & ~48.44\textsuperscript{\dag} & ~.8822\textsuperscript{\dag} & ~~~2.96\textsuperscript{\dag\dag} & -2.08 \\
 &  & \textsc{Quit} ($w=12$) & 2.65 & .8332 & ~48.34\textsuperscript{\dag} & ~~~.8819\textsuperscript{\dag\dag} & ~~~2.96\textsuperscript{\dag\dag} & -2.11 \\
 &  & \textsc{Quit} ($w=14$) & 2.24 & ~.8338\textsuperscript{\dag} & ~48.28\textsuperscript{\dag} & ~~~.8816\textsuperscript{\dag\dag} & ~~~2.96\textsuperscript{\dag\dag} & ~-2.13\textsuperscript{\dag} \\
 &  & \textsc{Quit} ($w=16$) & 1.94 & ~.8342\textsuperscript{\dag} & ~48.19\textsuperscript{\dag} & ~~~.8816\textsuperscript{\dag\dag} & ~~~2.96\textsuperscript{\dag\dag} & ~-2.16\textsuperscript{\dag} \\
\bottomrule
\end{tabular}
\caption{Complete window-size sweep on WMT24 for the window sizes not shown in Table~\ref{tab:main_results_wmt24}, following its conventions.}\label{tab:sweep_wmt24}
\end{table*}

\begin{table*}[t]
\centering
\footnotesize
\setlength{\tabcolsep}{3.5pt}
\begin{tabular}{@{}lllcccccc@{}}
\toprule
\textbf{NMT Model} & \textbf{Reranker} & \textbf{Acceleration} & \textbf{Speedup $\uparrow$} & \textbf{Rerank. $\uparrow$} & \textbf{ChrF++ $\uparrow$} & \textbf{xCOMET $\uparrow$} & \textbf{MetricX $\downarrow$} & \textbf{GEMBA $\uparrow$} \\
\midrule
\multirow{14}{*}{Qwen3} & \multirow{8}{*}{MBR} & Unaccelerated & 1.00 & .7345 & 34.06 & .3696 & 10.86 & -18.11 \\
 &  & PruneMBR & 1.02 & ~~~.7344\textsuperscript{\dag\dag} & ~~~34.04\textsuperscript{\dag\dag} & ~~~.3692\textsuperscript{\dag\dag} & ~~~10.86\textsuperscript{\dag\dag} & ~~~-18.10\textsuperscript{\dag\dag} \\
 &  & PMBR & 1.01 & ~~~.7339\textsuperscript{\dag\dag} & ~~~34.07\textsuperscript{\dag\dag} & ~~~.3689\textsuperscript{\dag\dag} & ~~~10.88\textsuperscript{\dag\dag} & ~~~-18.16\textsuperscript{\dag\dag} \\
 &  & \textsc{Quit} ($w=2$) & 8.23 & .7192 & ~33.73\textsuperscript{\dag} & .3531 & 11.31 & -18.71 \\
 &  & \textsc{Quit} ($w=4$) & 2.81 & ~.7280\textsuperscript{\dag} & ~~~33.89\textsuperscript{\dag\dag} & ~.3621\textsuperscript{\dag} & ~11.07\textsuperscript{\dag} & ~-18.36\textsuperscript{\dag} \\
 &  & \textsc{Quit} ($w=6$) & 1.83 & ~.7309\textsuperscript{\dag} & ~~~33.93\textsuperscript{\dag\dag} & ~.3654\textsuperscript{\dag} & ~10.98\textsuperscript{\dag} & ~-18.29\textsuperscript{\dag} \\
 &  & \textsc{Quit} ($w=8$) & 1.47 & ~~~.7322\textsuperscript{\dag\dag} & ~~~33.98\textsuperscript{\dag\dag} & ~~~.3671\textsuperscript{\dag\dag} & ~~~10.95\textsuperscript{\dag\dag} & ~~~-18.19\textsuperscript{\dag\dag} \\
 &  & \textsc{Quit} ($w=10$) & 1.30 & ~~~.7333\textsuperscript{\dag\dag} & ~~~34.00\textsuperscript{\dag\dag} & ~~~.3672\textsuperscript{\dag\dag} & ~~~10.92\textsuperscript{\dag\dag} & ~~~-18.20\textsuperscript{\dag\dag} \\
 &  & \textsc{Quit} ($w=12$) & 1.20 & ~~~.7336\textsuperscript{\dag\dag} & ~~~34.01\textsuperscript{\dag\dag} & ~~~.3683\textsuperscript{\dag\dag} & ~~~10.90\textsuperscript{\dag\dag} & ~~~-18.19\textsuperscript{\dag\dag} \\
 &  & \textsc{Quit} ($w=14$) & 1.14 & ~~~.7339\textsuperscript{\dag\dag} & ~~~34.04\textsuperscript{\dag\dag} & ~~~.3686\textsuperscript{\dag\dag} & ~~~10.88\textsuperscript{\dag\dag} & ~~~-18.16\textsuperscript{\dag\dag} \\
 &  & \textsc{Quit} ($w=16$) & 1.10 & ~~~.7340\textsuperscript{\dag\dag} & ~~~34.04\textsuperscript{\dag\dag} & ~~~.3689\textsuperscript{\dag\dag} & ~~~10.88\textsuperscript{\dag\dag} & ~~~-18.17\textsuperscript{\dag\dag} \\
\cmidrule{2-9}
 & \multirow{6}{*}{QE} & Unaccelerated & 1.00 & .7106 & 30.32 & .3705 & 10.10 & -18.24 \\
 &  & \textsc{Quit} ($w=2$) & 19.77 & .6675 & 32.17 & .3453 & 11.27 & -18.97 \\
 &  & \textsc{Quit} ($w=4$) & 7.75 & .6818 & 31.81 & .3532 & 10.88 & -18.60 \\
 &  & \textsc{Quit} ($w=6$) & 4.79 & .6888 & 31.59 & .3584 & 10.69 & ~-18.48\textsuperscript{\dag} \\
 &  & \textsc{Quit} ($w=8$) & 3.43 & .6935 & 31.28 & ~.3613\textsuperscript{\dag} & 10.54 & ~-18.37\textsuperscript{\dag} \\
 &  & \textsc{Quit} ($w=10$) & 2.68 & .6967 & 31.22 & ~.3626\textsuperscript{\dag} & 10.46 & ~-18.34\textsuperscript{\dag} \\
 &  & \textsc{Quit} ($w=12$) & 2.19 & .6996 & ~31.05\textsuperscript{\dag} & ~.3646\textsuperscript{\dag} & 10.38 & ~~~-18.30\textsuperscript{\dag\dag} \\
 &  & \textsc{Quit} ($w=14$) & 1.87 & .7021 & ~30.94\textsuperscript{\dag} & ~.3658\textsuperscript{\dag} & ~10.31\textsuperscript{\dag} & ~~~-18.24\textsuperscript{\dag\dag} \\
 &  & \textsc{Quit} ($w=16$) & 1.64 & .7036 & ~30.80\textsuperscript{\dag} & ~.3665\textsuperscript{\dag} & ~10.26\textsuperscript{\dag} & ~~~-18.22\textsuperscript{\dag\dag} \\
\midrule
\multirow{14}{*}{TranslateGemma} & \multirow{8}{*}{MBR} & Unaccelerated & 1.00 & .8129 & 36.25 & .5731 & 6.06 & -6.74 \\
 &  & PruneMBR & 1.01 & ~~~.8129\textsuperscript{\dag\dag} & ~~~36.25\textsuperscript{\dag\dag} & ~~~.5735\textsuperscript{\dag\dag} & ~~~6.06\textsuperscript{\dag\dag} & ~~~-6.71\textsuperscript{\dag\dag} \\
 &  & PMBR & 1.01 & ~~~.8125\textsuperscript{\dag\dag} & ~~~36.22\textsuperscript{\dag\dag} & ~~~.5717\textsuperscript{\dag\dag} & ~~~6.08\textsuperscript{\dag\dag} & ~-6.85\textsuperscript{\dag} \\
 &  & \textsc{Quit} ($w=2$) & 12.54 & .8074 & ~~~36.07\textsuperscript{\dag\dag} & ~.5633\textsuperscript{\dag} & 6.22 & -7.16 \\
 &  & \textsc{Quit} ($w=4$) & 4.76 & ~.8099\textsuperscript{\dag} & ~~~36.13\textsuperscript{\dag\dag} & ~.5673\textsuperscript{\dag} & ~6.17\textsuperscript{\dag} & ~-6.94\textsuperscript{\dag} \\
 &  & \textsc{Quit} ($w=6$) & 3.30 & ~.8108\textsuperscript{\dag} & ~~~36.15\textsuperscript{\dag\dag} & ~.5683\textsuperscript{\dag} & ~6.14\textsuperscript{\dag} & ~-6.91\textsuperscript{\dag} \\
 &  & \textsc{Quit} ($w=8$) & 2.61 & ~~~.8112\textsuperscript{\dag\dag} & ~~~36.18\textsuperscript{\dag\dag} & ~.5690\textsuperscript{\dag} & ~6.12\textsuperscript{\dag} & ~-6.87\textsuperscript{\dag} \\
 &  & \textsc{Quit} ($w=10$) & 2.18 & ~~~.8116\textsuperscript{\dag\dag} & ~~~36.17\textsuperscript{\dag\dag} & ~~~.5697\textsuperscript{\dag\dag} & ~~~6.10\textsuperscript{\dag\dag} & ~-6.82\textsuperscript{\dag} \\
 &  & \textsc{Quit} ($w=12$) & 1.89 & ~~~.8119\textsuperscript{\dag\dag} & ~~~36.16\textsuperscript{\dag\dag} & ~~~.5704\textsuperscript{\dag\dag} & ~~~6.10\textsuperscript{\dag\dag} & ~~~-6.80\textsuperscript{\dag\dag} \\
 &  & \textsc{Quit} ($w=14$) & 1.67 & ~~~.8121\textsuperscript{\dag\dag} & ~~~36.19\textsuperscript{\dag\dag} & ~~~.5711\textsuperscript{\dag\dag} & ~~~6.09\textsuperscript{\dag\dag} & ~~~-6.82\textsuperscript{\dag\dag} \\
 &  & \textsc{Quit} ($w=16$) & 1.51 & ~~~.8123\textsuperscript{\dag\dag} & ~~~36.19\textsuperscript{\dag\dag} & ~~~.5713\textsuperscript{\dag\dag} & ~~~6.10\textsuperscript{\dag\dag} & ~~~-6.79\textsuperscript{\dag\dag} \\
\cmidrule{2-9}
 & \multirow{6}{*}{QE} & Unaccelerated & 1.00 & .7721 & 36.17 & .5670 & 6.11 & -6.80 \\
 &  & \textsc{Quit} ($w=2$) & 20.40 & .7543 & ~~~36.08\textsuperscript{\dag\dag} & .5556 & 6.31 & -7.12 \\
 &  & \textsc{Quit} ($w=4$) & 8.09 & .7608 & ~~~36.15\textsuperscript{\dag\dag} & ~.5601\textsuperscript{\dag} & ~6.23\textsuperscript{\dag} & ~-7.03\textsuperscript{\dag} \\
 &  & \textsc{Quit} ($w=6$) & 4.97 & .7638 & ~~~36.16\textsuperscript{\dag\dag} & ~.5623\textsuperscript{\dag} & ~6.20\textsuperscript{\dag} & ~-7.02\textsuperscript{\dag} \\
 &  & \textsc{Quit} ($w=8$) & 3.55 & .7657 & ~~~36.16\textsuperscript{\dag\dag} & ~~~.5645\textsuperscript{\dag\dag} & ~6.18\textsuperscript{\dag} & ~-6.86\textsuperscript{\dag} \\
 &  & \textsc{Quit} ($w=10$) & 2.78 & .7670 & ~~~36.15\textsuperscript{\dag\dag} & ~~~.5651\textsuperscript{\dag\dag} & ~6.17\textsuperscript{\dag} & ~-6.88\textsuperscript{\dag} \\
 &  & \textsc{Quit} ($w=12$) & 2.29 & .7680 & ~~~36.17\textsuperscript{\dag\dag} & ~~~.5651\textsuperscript{\dag\dag} & ~~~6.16\textsuperscript{\dag\dag} & ~~~-6.85\textsuperscript{\dag\dag} \\
 &  & \textsc{Quit} ($w=14$) & 1.94 & ~.7688\textsuperscript{\dag} & ~~~36.19\textsuperscript{\dag\dag} & ~~~.5656\textsuperscript{\dag\dag} & ~~~6.15\textsuperscript{\dag\dag} & ~~~-6.84\textsuperscript{\dag\dag} \\
 &  & \textsc{Quit} ($w=16$) & 1.69 & ~.7694\textsuperscript{\dag} & ~~~36.20\textsuperscript{\dag\dag} & ~~~.5654\textsuperscript{\dag\dag} & ~~~6.14\textsuperscript{\dag\dag} & ~~~-6.81\textsuperscript{\dag\dag} \\
\midrule
\multirow{14}{*}{Hy-MT2} & \multirow{8}{*}{MBR} & Unaccelerated & 1.00 & .8298 & 41.19 & .6281 & 5.62 & -4.25 \\
 &  & PruneMBR & 1.01 & ~~~.8298\textsuperscript{\dag\dag} & ~~~41.20\textsuperscript{\dag\dag} & ~~~.6278\textsuperscript{\dag\dag} & ~~~5.62\textsuperscript{\dag\dag} & ~~~-4.19\textsuperscript{\dag\dag} \\
 &  & PMBR & 1.01 & ~~~.8294\textsuperscript{\dag\dag} & ~~~41.19\textsuperscript{\dag\dag} & ~~~.6279\textsuperscript{\dag\dag} & ~~~5.63\textsuperscript{\dag\dag} & ~~~-4.23\textsuperscript{\dag\dag} \\
 &  & \textsc{Quit} ($w=2$) & 12.28 & ~.8257\textsuperscript{\dag} & ~~~41.04\textsuperscript{\dag\dag} & ~.6175\textsuperscript{\dag} & 5.76 & ~-4.46\textsuperscript{\dag} \\
 &  & \textsc{Quit} ($w=4$) & 4.46 & ~~~.8279\textsuperscript{\dag\dag} & ~~~41.19\textsuperscript{\dag\dag} & ~.6226\textsuperscript{\dag} & ~5.69\textsuperscript{\dag} & ~-4.38\textsuperscript{\dag} \\
 &  & \textsc{Quit} ($w=6$) & 3.03 & ~~~.8285\textsuperscript{\dag\dag} & ~~~41.17\textsuperscript{\dag\dag} & ~~~.6245\textsuperscript{\dag\dag} & ~~~5.67\textsuperscript{\dag\dag} & ~-4.35\textsuperscript{\dag} \\
 &  & \textsc{Quit} ($w=8$) & 2.36 & ~~~.8289\textsuperscript{\dag\dag} & ~~~41.18\textsuperscript{\dag\dag} & ~~~.6262\textsuperscript{\dag\dag} & ~~~5.65\textsuperscript{\dag\dag} & ~~~-4.25\textsuperscript{\dag\dag} \\
 &  & \textsc{Quit} ($w=10$) & 1.97 & ~~~.8291\textsuperscript{\dag\dag} & ~~~41.15\textsuperscript{\dag\dag} & ~~~.6263\textsuperscript{\dag\dag} & ~~~5.64\textsuperscript{\dag\dag} & ~~~-4.29\textsuperscript{\dag\dag} \\
 &  & \textsc{Quit} ($w=12$) & 1.72 & ~~~.8292\textsuperscript{\dag\dag} & ~~~41.16\textsuperscript{\dag\dag} & ~~~.6278\textsuperscript{\dag\dag} & ~~~5.64\textsuperscript{\dag\dag} & ~~~-4.25\textsuperscript{\dag\dag} \\
 &  & \textsc{Quit} ($w=14$) & 1.53 & ~~~.8294\textsuperscript{\dag\dag} & ~~~41.17\textsuperscript{\dag\dag} & ~~~.6275\textsuperscript{\dag\dag} & ~~~5.64\textsuperscript{\dag\dag} & ~~~-4.22\textsuperscript{\dag\dag} \\
 &  & \textsc{Quit} ($w=16$) & 1.40 & ~~~.8295\textsuperscript{\dag\dag} & ~~~41.17\textsuperscript{\dag\dag} & ~~~.6279\textsuperscript{\dag\dag} & ~~~5.63\textsuperscript{\dag\dag} & ~~~-4.22\textsuperscript{\dag\dag} \\
\cmidrule{2-9}
 & \multirow{6}{*}{QE} & Unaccelerated & 1.00 & .7786 & 40.78 & .6219 & 5.71 & -4.88 \\
 &  & \textsc{Quit} ($w=2$) & 20.48 & .7618 & ~~~40.98\textsuperscript{\dag\dag} & .6105 & 5.87 & ~-4.77\textsuperscript{\dag} \\
 &  & \textsc{Quit} ($w=4$) & 8.12 & .7678 & ~~~40.89\textsuperscript{\dag\dag} & ~.6151\textsuperscript{\dag} & ~5.81\textsuperscript{\dag} & ~-4.65\textsuperscript{\dag} \\
 &  & \textsc{Quit} ($w=6$) & 4.95 & .7708 & ~~~40.90\textsuperscript{\dag\dag} & ~.6171\textsuperscript{\dag} & ~5.79\textsuperscript{\dag} & ~-4.65\textsuperscript{\dag} \\
 &  & \textsc{Quit} ($w=8$) & 3.59 & .7725 & ~~~40.90\textsuperscript{\dag\dag} & ~.6184\textsuperscript{\dag} & ~5.77\textsuperscript{\dag} & ~-4.74\textsuperscript{\dag} \\
 &  & \textsc{Quit} ($w=10$) & 2.80 & .7737 & ~~~40.90\textsuperscript{\dag\dag} & ~~~.6195\textsuperscript{\dag\dag} & ~5.76\textsuperscript{\dag} & ~-4.78\textsuperscript{\dag} \\
 &  & \textsc{Quit} ($w=12$) & 2.29 & .7747 & ~~~40.88\textsuperscript{\dag\dag} & ~~~.6197\textsuperscript{\dag\dag} & ~~~5.75\textsuperscript{\dag\dag} & ~-4.79\textsuperscript{\dag} \\
 &  & \textsc{Quit} ($w=14$) & 1.94 & ~.7755\textsuperscript{\dag} & ~~~40.86\textsuperscript{\dag\dag} & ~~~.6207\textsuperscript{\dag\dag} & ~~~5.75\textsuperscript{\dag\dag} & ~-4.79\textsuperscript{\dag} \\
 &  & \textsc{Quit} ($w=16$) & 1.70 & ~.7761\textsuperscript{\dag} & ~~~40.84\textsuperscript{\dag\dag} & ~~~.6210\textsuperscript{\dag\dag} & ~~~5.74\textsuperscript{\dag\dag} & ~~~-4.83\textsuperscript{\dag\dag} \\
\bottomrule
\end{tabular}
\caption{Complete window-size sweep on WMT25 for the window sizes not shown in Table~\ref{tab:main_results_wmt25}, following its conventions.}\label{tab:sweep_wmt25}
\end{table*}

\newcommand{\dagmark}{\rlap{\textsuperscript{\dag}}}
\newcommand{\ddagmark}{\rlap{\textsuperscript{\dag\dag}}}

\begin{table*}[t]
\centering
\footnotesize
\setlength{\tabcolsep}{4pt}
\begin{tabular}{@{}lllrcllll@{}}
\toprule
\textbf{Dataset}
& \textbf{NMT Model}
& \textbf{Window}
& \textbf{$\bar N$}
& \textbf{Speedup $\uparrow$}
& \textbf{ChrF++ $\uparrow$}
& \textbf{xCOMET $\uparrow$}
& \textbf{MetricX $\downarrow$}
& \textbf{GEMBA $\uparrow$} \\
\midrule

\multirow{27}{*}{WMT24}
& \multirow{9}{*}{Qwen3}
& Unaccelerated & 512.00 & 1.00 & 42.48 & .6998 & 6.21 & -9.96 \\
& & $w=2$  & 84.25 & 8.11 & 42.15\ddagmark & .6968\ddagmark & 6.28\ddagmark & -10.16\dagmark \\
& & $w=4$  & 268.35 & 2.48 & 42.36\ddagmark & .6985\ddagmark & 6.24\ddagmark & -10.02\ddagmark \\
& & $w=6$  & 374.27 & 1.63 & 42.42\ddagmark & .6992\ddagmark & 6.22\ddagmark & -9.96\ddagmark \\
& & $w=8$  & 430.57 & 1.33 & 42.44\ddagmark & .6989\ddagmark & 6.23\ddagmark & -9.98\ddagmark \\
& & $w=10$ & 460.37 & 1.19 & 42.46\ddagmark & .6988\ddagmark & 6.22\ddagmark & -9.98\ddagmark \\
& & $w=12$ & 477.99 & 1.12 & 42.47\ddagmark & .6993\ddagmark & 6.22\ddagmark & -9.98\ddagmark \\
& & $w=14$ & 488.04 & 1.07 & 42.47\ddagmark & .6996\ddagmark & 6.22\ddagmark & -9.97\ddagmark \\
& & $w=16$ & 496.39 & 1.04 & 42.48\ddagmark & .6997\ddagmark & 6.21\ddagmark & -9.96\ddagmark \\
\cmidrule{2-9}

& \multirow{9}{*}{TranslateGemma}
& Unaccelerated & 512.00 & 1.00 & 45.08 & .8341 & 3.42 & -3.60 \\
& & $w=2$  & 75.68 & 8.88 & 44.96\ddagmark & .8336\ddagmark & 3.44\ddagmark & -3.64\ddagmark \\
& & $w=4$  & 237.32 & 2.81 & 45.04\ddagmark & .8337\ddagmark & 3.42\ddagmark & -3.64\ddagmark \\
& & $w=6$  & 339.77 & 1.83 & 45.06\ddagmark & .8340\ddagmark & 3.42\ddagmark & -3.61\ddagmark \\
& & $w=8$  & 398.79 & 1.48 & 45.07\ddagmark & .8337\ddagmark & 3.42\ddagmark & -3.61\ddagmark \\
& & $w=10$ & 434.86 & 1.30 & 45.08\ddagmark & .8340\ddagmark & 3.41\ddagmark & -3.60\ddagmark \\
& & $w=12$ & 457.26 & 1.20 & 45.08\ddagmark & .8339\ddagmark & 3.42\ddagmark & -3.59\ddagmark \\
& & $w=14$ & 472.65 & 1.13 & 45.08\ddagmark & .8341\ddagmark & 3.42\ddagmark & -3.61\ddagmark \\
& & $w=16$ & 484.90 & 1.08 & 45.08\ddagmark & .8341\ddagmark & 3.42\ddagmark & -3.60\ddagmark \\
\cmidrule{2-9}

& \multirow{9}{*}{Hy-MT2}
& Unaccelerated & 512.00 & 1.00 & 50.14 & .8777 & 3.09 & -1.99 \\
& & $w=2$  & 65.74 & 9.34 & 50.07\ddagmark & .8774\ddagmark & 3.11\ddagmark & -2.05\dagmark \\
& & $w=4$  & 205.69 & 3.08 & 50.12\ddagmark & .8775\ddagmark & 3.09\ddagmark & -1.99\ddagmark \\
& & $w=6$  & 301.00 & 1.99 & 50.14\ddagmark & .8777\ddagmark & 3.09\ddagmark & -2.00\ddagmark \\
& & $w=8$  & 362.47 & 1.57 & 50.14\ddagmark & .8777\ddagmark & 3.09\ddagmark & -2.00\ddagmark \\
& & $w=10$ & 403.05 & 1.35 & 50.14\ddagmark & .8777\ddagmark & 3.09\ddagmark & -2.00\ddagmark \\
& & $w=12$ & 430.88 & 1.23 & 50.14\ddagmark & .8776\ddagmark & 3.09\ddagmark & -1.98\ddagmark \\
& & $w=14$ & 450.16 & 1.15 & 50.14\ddagmark & .8778\ddagmark & 3.09\ddagmark & -2.00\ddagmark \\
& & $w=16$ & 465.10 & 1.09 & 50.14\ddagmark & .8777\ddagmark & 3.09\ddagmark & -1.98\ddagmark \\

\midrule

\multirow{27}{*}{WMT25}
& \multirow{9}{*}{Qwen3}
& Unaccelerated & 512.00 & 1.00 & 34.85 & .3238 & 12.63 & -19.80 \\
& & $w=2$  & 79.76 & 6.76 & 34.43\dagmark & .3192\dagmark & 12.76\dagmark & -20.01\dagmark \\
& & $w=4$  & 242.09 & 2.19 & 34.70\ddagmark & .3213\ddagmark & 12.68\ddagmark & -19.98\dagmark \\
& & $w=6$  & 341.36 & 1.54 & 34.77\ddagmark & .3225\ddagmark & 12.64\ddagmark & -19.87\ddagmark \\
& & $w=8$  & 399.55 & 1.31 & 34.80\ddagmark & .3232\ddagmark & 12.65\ddagmark & -19.87\ddagmark \\
& & $w=10$ & 436.95 & 1.19 & 34.81\ddagmark & .3229\ddagmark & 12.65\ddagmark & -19.88\ddagmark \\
& & $w=12$ & 461.87 & 1.12 & 34.83\ddagmark & .3232\ddagmark & 12.64\ddagmark & -19.84\ddagmark \\
& & $w=14$ & 477.64 & 1.08 & 34.81\ddagmark & .3235\ddagmark & 12.64\ddagmark & -19.84\ddagmark \\
& & $w=16$ & 491.28 & 1.05 & 34.84\ddagmark & .3236\ddagmark & 12.64\ddagmark & -19.79\ddagmark \\
\cmidrule{2-9}

& \multirow{9}{*}{TranslateGemma}
& Unaccelerated & 512.00 & 1.00 & 36.81 & .5397 & 6.64 & -7.48 \\
& & $w=2$  & 59.85 & 9.43 & 36.64\ddagmark & .5371\ddagmark & 6.69\ddagmark & -7.49\ddagmark \\
& & $w=4$  & 169.09 & 3.40 & 36.70\ddagmark & .5363\dagmark & 6.66\ddagmark & -7.54\dagmark \\
& & $w=6$  & 249.41 & 2.29 & 36.74\ddagmark & .5372\ddagmark & 6.66\ddagmark & -7.60\dagmark \\
& & $w=8$  & 311.28 & 1.81 & 36.77\ddagmark & .5367\ddagmark & 6.67\ddagmark & -7.57\dagmark \\
& & $w=10$ & 359.31 & 1.54 & 36.77\ddagmark & .5375\ddagmark & 6.66\ddagmark & -7.55\ddagmark \\
& & $w=12$ & 397.55 & 1.37 & 36.78\ddagmark & .5382\ddagmark & 6.65\ddagmark & -7.53\ddagmark \\
& & $w=14$ & 427.25 & 1.26 & 36.77\ddagmark & .5378\ddagmark & 6.65\ddagmark & -7.54\ddagmark \\
& & $w=16$ & 452.50 & 1.17 & 36.80\ddagmark & .5389\ddagmark & 6.64\ddagmark & -7.54\ddagmark \\
\cmidrule{2-9}

& \multirow{9}{*}{Hy-MT2}
& Unaccelerated & 512.00 & 1.00 & 42.00 & .5994 & 6.10 & -4.91 \\
& & $w=2$  & 60.23 & 9.24 & 41.74\ddagmark & .5945\dagmark & 6.23\dagmark & -5.32 \\
& & $w=4$  & 168.03 & 3.37 & 41.87\ddagmark & .5969\ddagmark & 6.15\dagmark & -5.15 \\
& & $w=6$  & 252.65 & 2.22 & 41.93\ddagmark & .5973\ddagmark & 6.12\ddagmark & -5.06\dagmark \\
& & $w=8$  & 316.98 & 1.75 & 41.96\ddagmark & .5980\ddagmark & 6.12\ddagmark & -5.01\dagmark \\
& & $w=10$ & 368.15 & 1.48 & 41.98\ddagmark & .5978\ddagmark & 6.11\ddagmark & -4.98\dagmark \\
& & $w=12$ & 409.52 & 1.31 & 41.97\ddagmark & .5987\ddagmark & 6.11\ddagmark & -4.95\ddagmark \\
& & $w=14$ & 438.35 & 1.21 & 41.98\ddagmark & .5985\ddagmark & 6.11\ddagmark & -4.95\ddagmark \\
& & $w=16$ & 464.20 & 1.13 & 41.99\ddagmark & .5988\ddagmark & 6.11\ddagmark & -4.90\ddagmark \\

\bottomrule
\end{tabular}

\caption{Translation quality when using ChrF as the utility function for MBR decoding, with a maximum candidate budget of $N_{\max}=512$. Results are reported for the unaccelerated baseline and \textsc{Quit} across window sizes $w\in\{2,4,6,8,10,12,14,16\}$ on WMT24 and WMT25. Mean $N$ denotes the average number of candidates used per sentence. Speedup is estimated as the unaccelerated generation-plus-utility time divided by the corresponding accelerated time. Equivalence markers follow the conventions of the main results.}

\label{tab:window_quality_results}
\end{table*}

\clearpage

\begin{table*}[t]
\centering
\small
\setlength{\tabcolsep}{5pt}
\begin{tabular}{@{}llcc@{}}
\toprule
\textbf{Test set} & \textbf{NMT model} & \textbf{MBR} & \textbf{QE} \\
\midrule
WMT24 & Qwen3 & 0.721 & 0.278 \\
WMT24 & TranslateGemma & 0.749 & 0.242 \\
WMT24 & Hy-MT2 & 0.784 & 0.192 \\
WMT25 & Qwen3 & 0.663 & 0.269 \\
WMT25 & TranslateGemma & 0.705 & 0.288 \\
WMT25 & Hy-MT2 & 0.734 & 0.286 \\
\midrule
\multicolumn{2}{l}{\textbf{Average}} & \textbf{0.726} & \textbf{0.259} \\
\bottomrule
\end{tabular}
\caption{Pooled PRR with uncertainty $\Delta_b^{R}(s)$ and proxy early-stopping risk $\mathcal{L}_{bw}^{\mathrm{proxy}}(s)$ (Equation~\ref{eq:proxy_early_stopping_risk}), $w=8$ (windows $b=1,\ldots,7$). The full-data evaluation and window construction match Table~\ref{tab:prr_w8}; the risk is computed from reranking scores. Higher PRR is better.}
\label{tab:prr_proxy_w8}
\end{table*}

\begin{table*}[t]
\centering
\small
\setlength{\tabcolsep}{5pt}
\begin{tabular}{@{}llcccc@{}}
\toprule
\multirow{2}{*}{\textbf{Test set}} &
\multirow{2}{*}{\textbf{NMT model}} &
\multicolumn{2}{c}{\textbf{MBR}} &
\multicolumn{2}{c}{\textbf{QE}} \\
\cmidrule(lr){3-4}\cmidrule(lr){5-6}
& & \textbf{xCOMET} & \textbf{GEMBA} & \textbf{xCOMET} & \textbf{GEMBA} \\
\midrule
WMT24 & Qwen3 & 0.416 & 0.399 & 0.106 & 0.089 \\
WMT24 & TranslateGemma & 0.387 & 0.390 & 0.110 & 0.103 \\
WMT24 & Hy-MT2 & 0.491 & 0.556 & 0.103 & 0.090 \\
WMT25 & Qwen3 & 0.267 & 0.107 & 0.109 & 0.088 \\
WMT25 & TranslateGemma & 0.286 & 0.331 & 0.091 & 0.080 \\
WMT25 & Hy-MT2 & 0.318 & 0.484 & 0.104 & 0.098 \\
\midrule
\multicolumn{2}{l}{\textbf{Average}} & \textbf{0.361} & \textbf{0.378} & \textbf{0.104} & \textbf{0.091} \\
\bottomrule
\end{tabular}
\caption{Pooled PRR with uncertainty $\Delta_b^{R}(s)$ and early-stopping risk $\mathcal{L}_{bw}(s)$ (Equation~\ref{eq:early_stopping_risk}), $w=4$ (windows $b=1,\ldots,15$). Higher PRR is better.}
\label{tab:prr_w4}
\end{table*}

\begin{table*}[t]
\centering
\small
\setlength{\tabcolsep}{5pt}
\begin{tabular}{@{}llcccc@{}}
\toprule
\multirow{2}{*}{\textbf{Test set}} &
\multirow{2}{*}{\textbf{NMT model}} &
\multicolumn{2}{c}{\textbf{MBR}} &
\multicolumn{2}{c}{\textbf{QE}} \\
\cmidrule(lr){3-4}\cmidrule(lr){5-6}
& & \textbf{xCOMET} & \textbf{GEMBA} & \textbf{xCOMET} & \textbf{GEMBA} \\
\midrule
WMT24 & Qwen3 & 0.389 & 0.377 & 0.246 & 0.211 \\
WMT24 & TranslateGemma & 0.357 & 0.372 & 0.283 & 0.271 \\
WMT24 & Hy-MT2 & 0.490 & 0.557 & 0.256 & 0.237 \\
WMT25 & Qwen3 & 0.197 & 0.065 & 0.208 & 0.160 \\
WMT25 & TranslateGemma & 0.216 & 0.271 & 0.184 & 0.183 \\
WMT25 & Hy-MT2 & 0.258 & 0.452 & 0.222 & 0.255 \\
\midrule
\multicolumn{2}{l}{\textbf{Average}} & \textbf{0.318} & \textbf{0.349} & \textbf{0.233} & \textbf{0.220} \\
\bottomrule
\end{tabular}
\caption{Pooled PRR with uncertainty $\Delta_b^{R}(s)$ and early-stopping risk $\mathcal{L}_{bw}(s)$ (Equation~\ref{eq:early_stopping_risk}), $w=16$ (windows $b=1,\ldots,3$). Higher PRR is better.}
\label{tab:prr_w16}
\end{table*}

\clearpage

\clearpage
\begin{table*}[t]
\centering
\small
\setlength{\tabcolsep}{3.5pt}
\begin{tabular}{@{}lcrrrrrr@{}}
\toprule
\multirow{2}{*}{\textbf{Method}} &
\multirow{2}{*}{$\boldsymbol{w}$} &
\multicolumn{2}{c}{\textbf{Qwen3}} &
\multicolumn{2}{c}{\textbf{TranslateGemma}} &
\multicolumn{2}{c}{\textbf{Hy-MT2}} \\
\cmidrule(lr){3-4}\cmidrule(lr){5-6}\cmidrule(lr){7-8}
& & \textbf{Generation} & \textbf{Reranking} & \textbf{Generation} & \textbf{Reranking} & \textbf{Generation} & \textbf{Reranking} \\
\midrule
\textsc{Quit} & 2 & 28,663.0 & 285.97 & 24,184.0 & 192.77 & 34,071.0 & 140.53 \\
\textsc{Quit} & 4 & 79,854.0 & 2,369.35 & 62,531.8 & 1,440.79 & 88,251.2 & 1,034.54 \\
\textsc{Quit} & 6 & 120,380 & 5,080.10 & 93,674.8 & 2,946.10 & 135,153 & 2,255.54 \\
\textsc{Quit} & 8 & 155,297 & 7,524.67 & 121,476 & 4,395.07 & 178,592 & 3,587.21 \\
\textsc{Quit} & 10 & 182,116 & 9,428.58 & 146,152 & 5,797.00 & 217,507 & 4,901.43 \\
\textsc{Quit} & 12 & 202,580 & 10,963.7 & 169,529 & 7,143.56 & 253,175 & 6,200.59 \\
\textsc{Quit} & 14 & 219,347 & 12,261.7 & 191,172 & 8,383.02 & 285,284 & 7,461.55 \\
\textsc{Quit} & 16 & 231,725 & 13,298.9 & 212,182 & 9,725.38 & 314,869 & 8,709.65 \\
\midrule
Unaccelerated MBR & $\infty$ & 266,352 & 16,803.0 & 317,548 & 16,803.0 & 467,484 & 16,803.0 \\
PMBR & -- & 266,352 & 7,301.66 & 317,548 & 6,079.76 & 467,484 & 5,497.37 \\
PruneMBR & -- & 266,352 & 2,978.32 & 317,548 & 4,977.67 & 467,484 & 8,002.77 \\
\bottomrule
\end{tabular}
\caption{Complete candidate-generation and MBR reranking wall-clock times (seconds) on WMT24 with a maximum budget of 512 candidates.}
\label{tab:runtime_wmt24}
\end{table*}

\begin{table*}[t]
\centering
\small
\setlength{\tabcolsep}{3.5pt}
\begin{tabular}{@{}lcrrrrrr@{}}
\toprule
\multirow{2}{*}{\textbf{Method}} &
\multirow{2}{*}{$\boldsymbol{w}$} &
\multicolumn{2}{c}{\textbf{Qwen3}} &
\multicolumn{2}{c}{\textbf{TranslateGemma}} &
\multicolumn{2}{c}{\textbf{Hy-MT2}} \\
\cmidrule(lr){3-4}\cmidrule(lr){5-6}\cmidrule(lr){7-8}
& & \textbf{Generation} & \textbf{Reranking} & \textbf{Generation} & \textbf{Reranking} & \textbf{Generation} & \textbf{Reranking} \\
\midrule
\textsc{Quit} & 2 & 53,678.5 & 138.07 & 41,504.1 & 73.95 & 63,015.5 & 76.06 \\
\textsc{Quit} & 4 & 156,733 & 1,131.16 & 109,052 & 564.03 & 173,237 & 621.09 \\
\textsc{Quit} & 6 & 240,220 & 2,502.81 & 157,107 & 1,028.57 & 254,613 & 1,150.87 \\
\textsc{Quit} & 8 & 297,719 & 3,641.88 & 197,999 & 1,478.28 & 326,227 & 1,706.51 \\
\textsc{Quit} & 10 & 336,032 & 4,464.99 & 237,644 & 2,001.33 & 391,669 & 2,293.50 \\
\textsc{Quit} & 12 & 363,642 & 5,136.86 & 273,125 & 2,525.07 & 447,977 & 2,862.44 \\
\textsc{Quit} & 14 & 382,797 & 5,627.80 & 308,877 & 3,095.56 & 502,860 & 3,494.68 \\
\textsc{Quit} & 16 & 397,191 & 6,044.74 & 340,941 & 3,678.19 & 550,413 & 4,092.32 \\
\midrule
Unaccelerated MBR & $\infty$ & 435,665 & 7,320.00 & 514,263 & 7,320.00 & 767,318 & 7,320.00 \\
PMBR & -- & 435,665 & 3,030.74 & 514,263 & 2,637.34 & 767,318 & 2,741.44 \\
PruneMBR & -- & 435,665 & 284.15 & 514,263 & 295.94 & 767,318 & 423.40 \\
\bottomrule
\end{tabular}
\caption{Complete candidate-generation and MBR reranking wall-clock times (seconds) on WMT25 with a maximum budget of 512 candidates.}
\label{tab:runtime_wmt25}
\end{table*}

\end{document}